\documentclass[letterpaper, 10 pt, conference]{ieeeconf}
\IEEEoverridecommandlockouts
\usepackage{times}
\usepackage{graphics}
\usepackage{graphicx}
\usepackage{subfigure}
\usepackage{amsmath,amssymb,amsopn,amstext,amsfonts}
\usepackage{cancel}
\usepackage[space]{cite}
\usepackage{pdfsync}
\usepackage{balance}
\usepackage{color}
\usepackage{mathtools}
\usepackage{bm}

\usepackage{diagbox}
\usepackage{float}
\usepackage{epstopdf}
\usepackage{pifont}
\usepackage{fixltx2e}
\usepackage{amsmath}
\usepackage{mathrsfs}
\usepackage{multirow}
\usepackage{url}
\usepackage{verbatim}
\usepackage{caption}
\usepackage{adjustbox}
\usepackage{booktabs}
\usepackage{threeparttable}
\usepackage[table]{xcolor}
\usepackage{makecell}
\usepackage{tabularx}
\usepackage{arydshln}
\usepackage[normalem]{ulem}
\usepackage{ifthen}
\usepackage{makecell}
\usepackage{algorithmicx}
\usepackage{array}
\usepackage{textcomp}
\usepackage{stfloats}
\usepackage{siunitx}
\definecolor{lightred}{rgb}{0.976470, 0.85490, 0.721568}
\definecolor{lightblue}{rgb}{0.68, 0.85, 0.9}
\definecolor{lightgreen}{rgb}{0.67, 0.88, 0.69}
\definecolor{lightyellow}{rgb}{1.0, 1.0, 0.6}

\makeatletter
\let\NAT@parse\undefined
\makeatother
\usepackage[linkcolor=red,citecolor=green,urlcolor=red,colorlinks=true]{hyperref}

\title{\LARGE \bf MemGS: Memory-Efficient Gaussian Splatting for Real-Time SLAM}
\author{Yinlong Bai, Hongxin Zhang, Sheng Zhong, Junkai Niu, Hai Li, Yijia He, Yi Zhou 
\thanks{Yinlong Bai, Hongxin Zhang, Sheng Zhong, Junkai Niu and Yi Zhou are with the Neuromorphic Automation and Intelligence Lab
(NAIL) at School of Robotics, Hunan University, Changsha, China. Email: \{yinlonga, zhx\_2514, bell, junkainiu, eeyzhou\}@hnu.edu.cn.}
\thanks{Hai Li and Yijia He are with the TCL RayNeo (RayNeo), Ningbo, China. Email: lihai@rayneo.com, heyijia2016@gmail.com}
\thanks{Corresponding author: Yi Zhou.}
\thanks{This work was supported by the National Key Research and Development Project of China under Grant 2023YFB4706600.}
}

\usepackage[linesnumbered,ruled,vlined]{algorithm2e}

\begin{document}
\maketitle
\thispagestyle{empty}
\pagestyle{empty}

\begin{abstract}
\label{sec:abstract}
Recent advancements in 3D Gaussian Splatting (3DGS) have made a significant impact on rendering and reconstruction techniques. 
{Current research predominantly focuses on improving rendering performance and reconstruction quality using high-performance desktop GPUs, largely overlooking applications for embedded platforms like micro air vehicles (MAVs). 
These devices, with their limited computational resources and memory, often face a trade-off between system performance and reconstruction quality.}
In this paper, we improve existing methods in terms of GPU memory usage while enhancing rendering quality.
{Specifically, to address redundant 3D Gaussian primitives in SLAM, we propose merging them in voxel space based on geometric similarity. 
This reduces GPU memory usage without impacting system runtime performance.}
{Furthermore, rendering quality is improved by initializing 3D Gaussian primitives via Patch-Grid (PG) point sampling, enabling more accurate modeling of the entire scene.}
Quantitative and qualitative evaluations on publicly available datasets demonstrate the effectiveness of our improvements.
\end{abstract}
\section*{Multimedia Material}
\label{sec:multimedia}
\phantomsection
\noindent Code: {\small \href{https://github.com/NAIL-HNU/MemGS_SLAM.git}{github.com/NAIL-HNU/MemGS\_SLAM.git} \label{code:github}}

\section{{Introduction}}
\label{sec: introduction}
Simultaneous Localization and Mapping (SLAM) is a fundamental problem in robotics and computer vision, aiming to estimate the trajectory of robots while
reconstructing the 3D structure of unknown environments. 
It plays an important role in Augmented Reality (AR) and Virtual Reality (VR) applications, which allow edge devices to estimate their 6-DoF poses for rendering virtual content in real scenes. 
{Traditional approaches, such as KinectFusion~\cite{kinectfusion}, BundleFusion~\cite{dai2017bundlefusion}, and InfiniTAM~\cite{infinitam}, utilize the Truncated Signed Distance Function (TSDF) on RGB-D data to achieve 3D surface reconstruction.}
{Other scene representations employed in dense visual SLAM include point clouds~\cite{6599048} and surfels~\cite{8794101}.}
{Recently, 3D Implicit Neural Representations (INR) and explicit differentiable rendering, such as Neural Radiance Fields~\cite{nerf} (NeRF) and 3D Gaussian Splatting~\cite{kerbl3Dgaussians} (3DGS), have gained widespread adoption for representing 3D objects and scenes.}
Consequently, recent SLAM research has seen the emergence of two major paradigms: NeRF-based SLAM~\cite{pointslam, sucar2021imap, zhu2022nice,rosinol2022nerf, maggio2022loc } and 3DGS-based SLAM~\cite{Keetha_2024_CVPR, Matsuki_2024_CVPR, ha2024rgbdgsicpslam, hhuang2024photoslam, peng2024rtgslam}.
\begin{figure}[tbp]
	\centering
	\subfigure[Initial Points\label{fig:initialization}]{\includegraphics[width=0.23\textwidth]{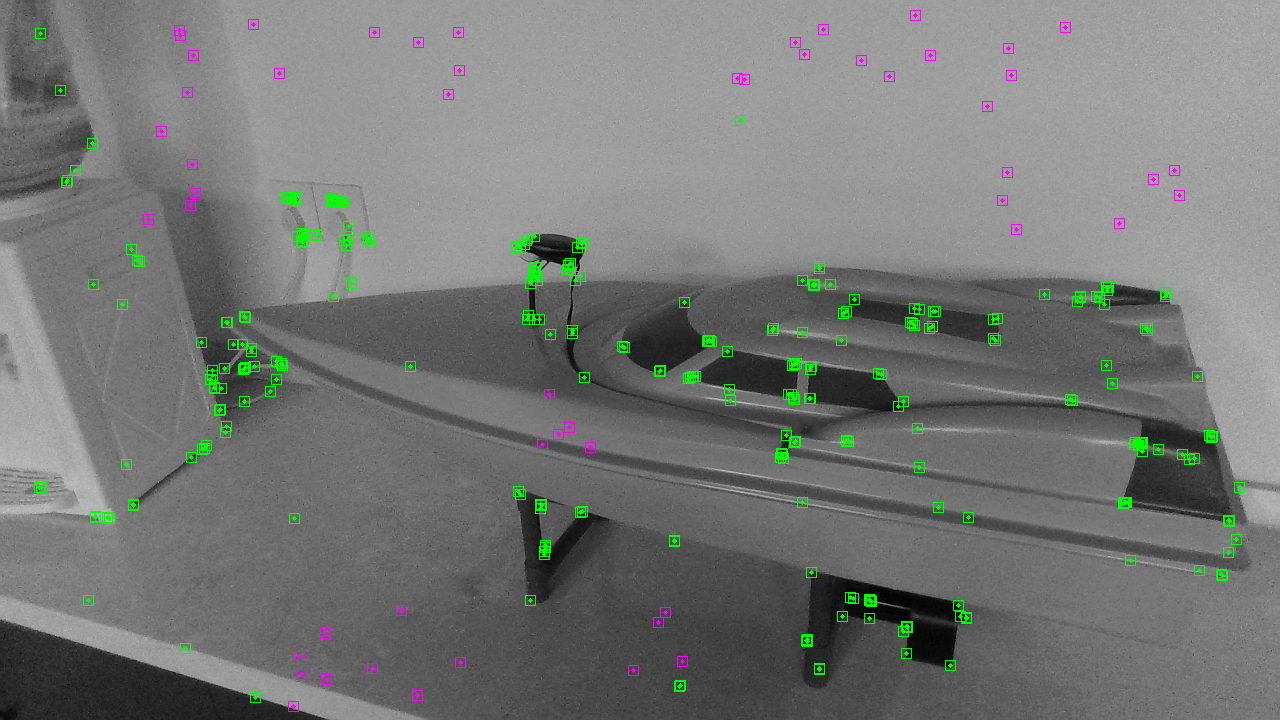}}\hspace{5pt}
	\subfigure[Densification\label{fig:densification}]{\includegraphics[width=0.23\textwidth]{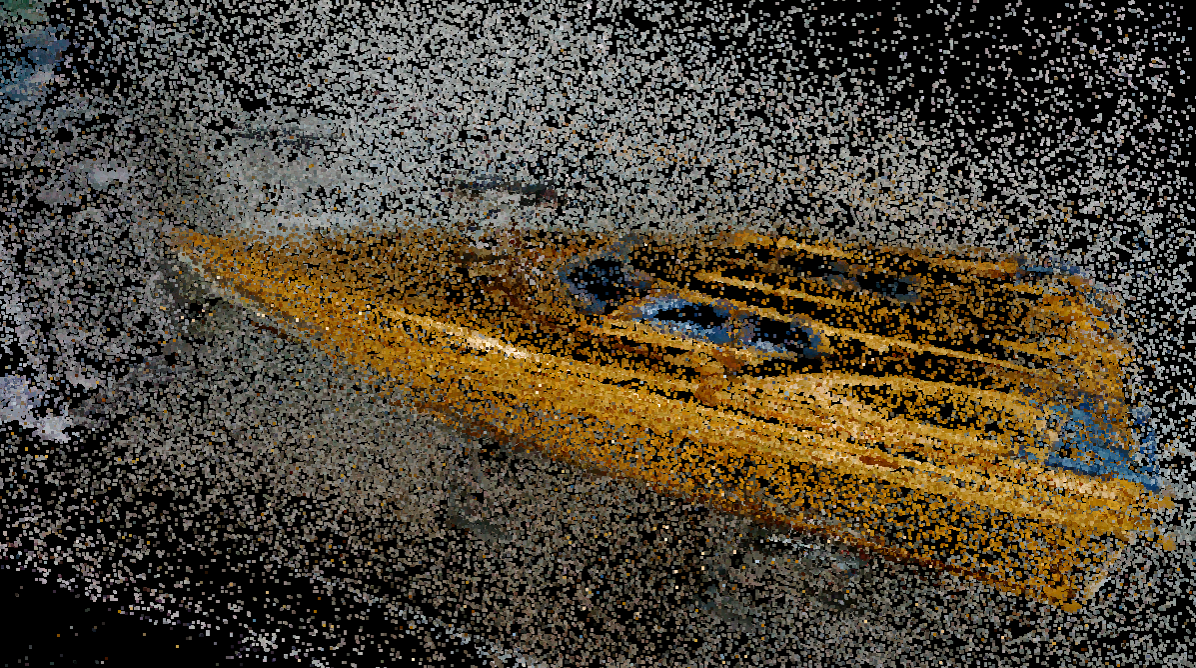}}\\
	\subfigure[3D Gaussian Mapping\label{fig:3d_gaussian_mapping}]{\includegraphics[width=0.23\textwidth]{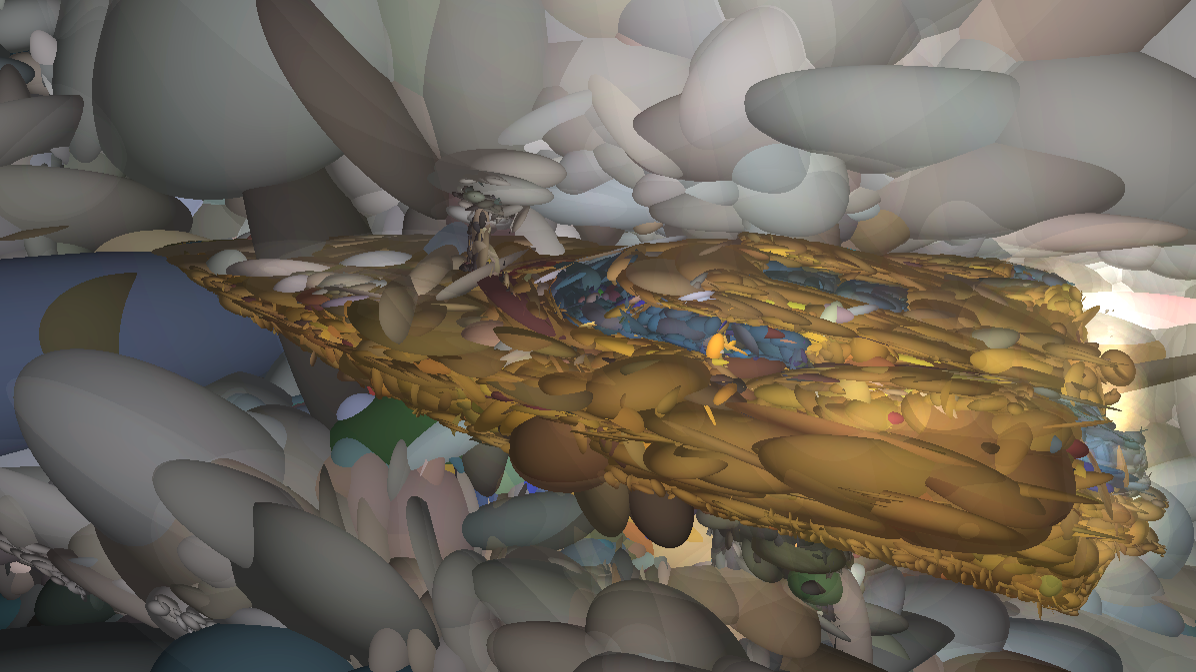}}\hspace{5pt}
	\subfigure[Online Reconstruction\label{fig:online_reconstruction}]{\includegraphics[width=0.23\textwidth]{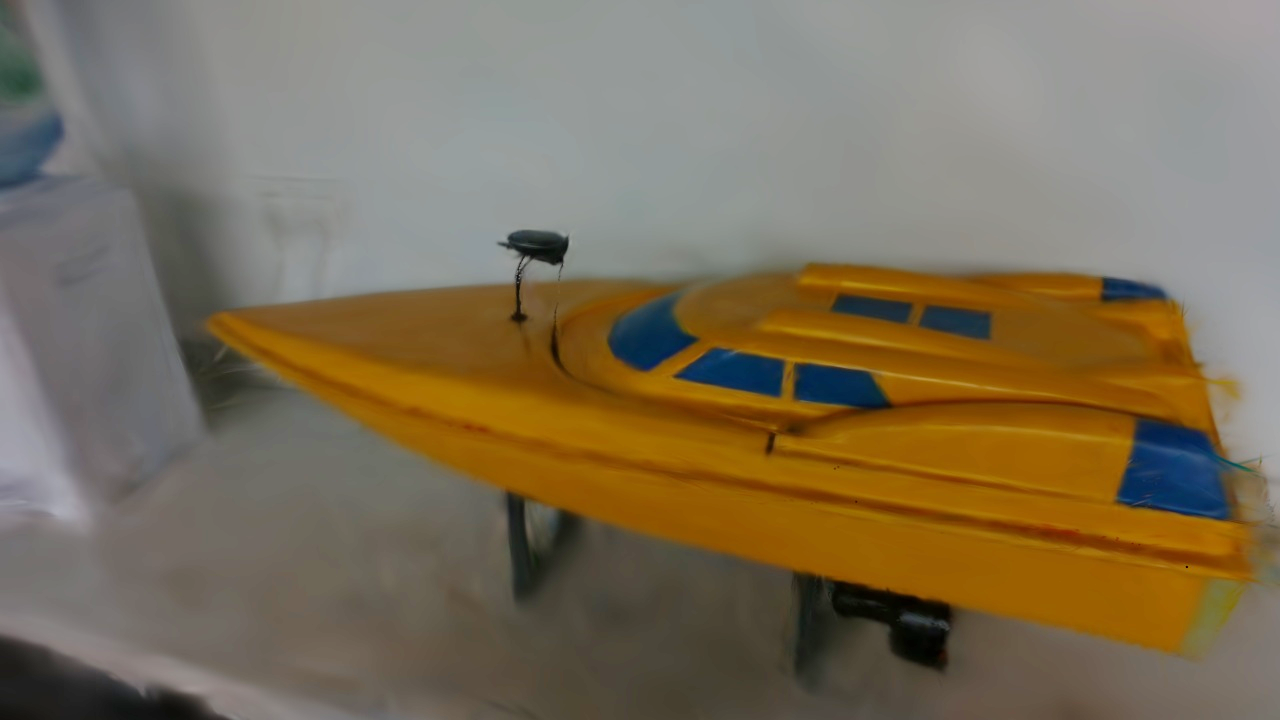}}
	\caption{Illustration of running the proposed system on real-world. (a) shows the keypoints (green) and sampling points (pink) on the image. (b) is the densification point clouds after initialization. (c) shows 3D Gaussian primitives of the scene. (d) shows the online reconstruction from camera viewpoint.}
	\vspace{-1.5em}
	\label{fig:eye_catcher}
\end{figure}

\textbf{NeRF-based SLAM.}
{The first method that incorporates NeRF into a SLAM pipeline is iMAP~\cite{sucar2021imap}, which employs a single Multi-Layer Perceptron (MLP) for environment reconstruction.}
Given the camera poses, iMAP~\cite{sucar2021imap} can render the scene from any viewpoint by minimizing the photometric error between the rendered and input images. 
{To reduce computational costs, NICE-SLAM~\cite{zhu2022nice} further improves scene representation by employing multi-resolution hybrid feature grids.}
ESLAM~\cite{eslam} and Co-SLAM~\cite{coslam} leverage the Instant-NGP~\cite{muller2022instant} and TensoRF~\cite{TensoRF}, respectively, to enhance mapping speed via axis-aligned feature planes and joint coordinate-parametric representations. 
However, a significant bottleneck of NeRF-based SLAM methods is the high computational cost and memory usage. 
Moreover, the training process usually takes hours or even days to converge, {making these methods impractical for real-time applications in real-world scenarios.} 

\textbf{3DGS-based SLAM.}
{Different from} NeRF-based SLAM methods, the differentiable rendering of 3DGS~\cite{kerbl3Dgaussians} via rasterization is more efficient, {and thus}, making it suitable for real-time SLAM. 
{3DGS-based SLAM methods model a scene using 3D Gaussian primitives, each characterized by attributes of orientation, scale, color, and opacity.}
Compared to feature-based SLAM~\cite{MurArtal15tro, campos2021orb3} or direct SLAM~\cite{7898369, 8593376, 9546534}, {however, 3DGS-based methods} have the disadvantage of slower convergence and cannot improve camera pose accuracy while fulfilling real-time requirements.
Existing 3DGS-based SLAM methods can be categorized into two {sub-categories}: coupled methods~\cite{Keetha_2024_CVPR,Matsuki_2024_CVPR,ha2024rgbdgsicpslam} and decoupled methods~\cite{hhuang2024photoslam, peng2024rtgslam}.
{The former can simultaneously optimize camera poses in the front-end and 3D Gaussian mapping in the back-end by sharing the same Gaussian map. Among them,
the covariances are shared between Generalized Iterative Closest Point~\cite{koide2021voxelized} (G-ICP) and 3D Gaussian primitives through scale-aligning techniques to foster faster convergence, as proposed in \cite{ha2024rgbdgsicpslam}.
}
{In contrast, the latter achieves faster speed and higher accuracy in initializing camera poses by incorporating ORB-SLAM3~\cite{campos2021orb3} or ICP~\cite{icp} algorithms as the front-end.}
{However, decoupled approaches~\cite{hhuang2024photoslam, peng2024rtgslam} heavily rely on an independent tracking thread to estimate poses and initialize 3D Gaussian primitives through feature extraction and matching, but these features are often more concentrated and sparse, limiting their ability to model the entire scene effectively}
{Additionally, 3DGS-based SLAM reveals that a significant portion of 3D Gaussian primitives exhibit similar geometry (\textit{i.e.}, position and covariance), as observed in~\cite{Lee_2024_CVPR}, resulting in redundant storage and slower optimization speeds.}

Hence, the goal of {this work} is to address the aforementioned limitations in existing 3DGS-based SLAM methods.
{In particular, we employ a 3D Gaussian representation to explore its potential for real-time SLAM, as illustrated in Fig.~\ref{fig:eye_catcher}. The proposed system can run on an edge device for online reconstruction. }

\textbf{\emph{Contributions:}}
\begin{itemize}
    \item {A novel initialization method for 3D Gaussian primitives, termed Patch-Grid (PG) sampling, is introduced to model the entire scene, significantly enhancing rendering quality.}
    \item {An enhanced approach to reduce GPU memory usage by merging redundant 3D Gaussian primitives with geometric similarity within the same voxel, without compromising runtime performance;}
    \item {Comprehensive evaluations on two commonly used datasets demonstrate the superior performance of our method, validating its suitability for real-time robotics applications.}
\end{itemize}

The rest of the paper is organized as follows.
First, we provide a discussion of our method, particularly focusing on the item listed in the contribution (Sec.~\ref{sec:Methodology}).
Then the experimental evaluation is provided in Sec.~\ref{sec:Experiments}, and finally the conclusion is made in Sec.~\ref{sec:conclusion}.
\section{{Methodology}}
\label{sec:Methodology}
In this section, we detail our proposed method. 
{First, we present a 3DGS-based SLAM system that incorporates ORB-SLAM3~\cite{campos2021orb3} as the front-end and 3D Gaussian mapping as the back-end.}
{Second, we provide a concise overview of the front-end's process for estimating and updating camera poses and map points, alongside the initialization of 3D Gaussian primitives using Patch-Grid (PG) sampling (Sec.~\ref{subsec:Patch-Grid}).}
{Third, we elaborate our proposed method for merging 3D Gaussian primitives with geometric similarity using the voxel map (Sec.~\ref{subsec:GS Splatting}).}
{Finally, we provide a detailed description of the mapping module, which jointly optimizes the 3D Gaussian map by constructing color and isotropic loss functions, along with depth loss (Sec.~\ref{subsec:Mapping}).}

\begin{figure*}[t]
    \centering
    \includegraphics[width=0.98\textwidth]{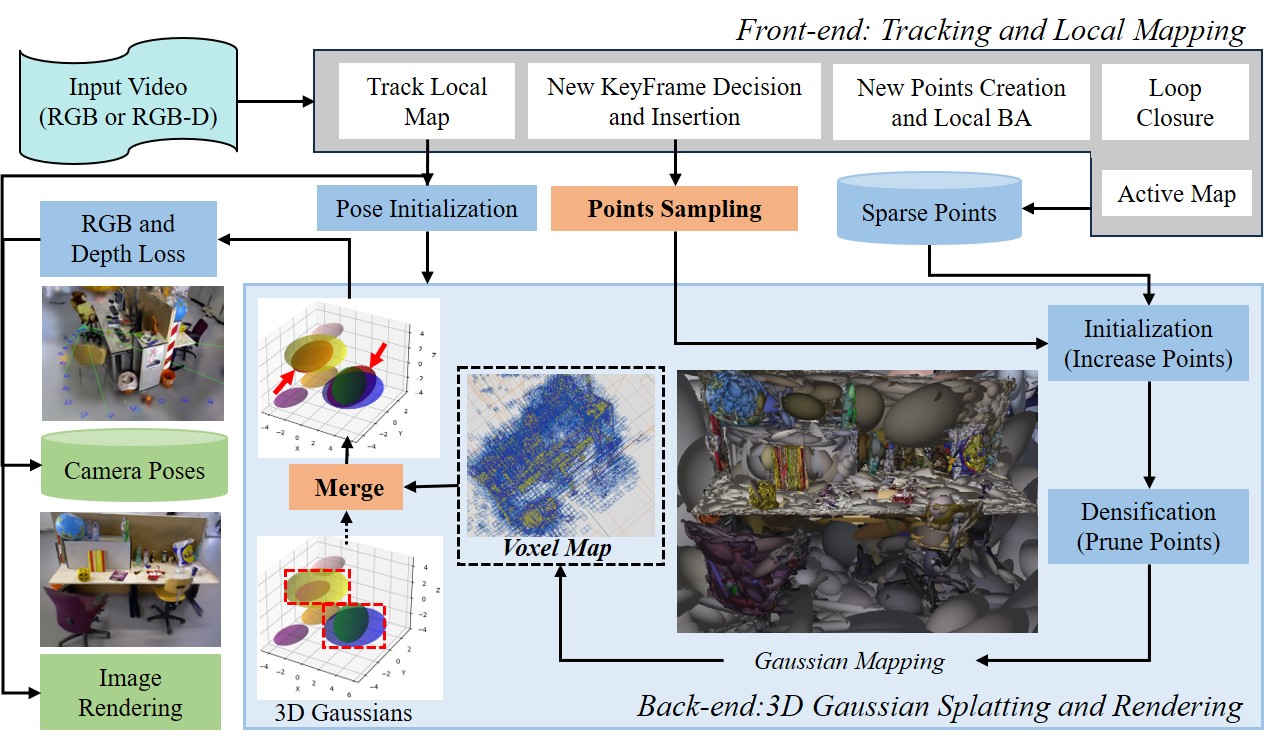}
    \caption{The proposed system framework and our work are highlighted in orange. In the front-end, We use ORB-SLAM3~\cite{campos2021orb3} as visual odometry to obtain the initial poses and sparse map points. Then we perform PG sampling for keyframes to initialize 3D Gaussian primitives. In the back-end, we use densification to generate dense point clouds, then merge geometrically similar 3D Gaussian primitives in the same voxel, and finally construct a joint loss function to model the entire scene.}
    \label{fig:system flowchart}
    \vspace{-1.5em}
\end{figure*}
\subsection{{Tracking}}
\label{subsec:Tracking}
In traditional SLAM~\cite{MurArtal15tro, campos2021orb3}, camera poses are usually estimated by minimizing the reprojection error between observed keypoints and their corresponding 3D points, 
which can be formulated as a non-linear optimization problem. In the local mapping thread, the camera poses and sparse map points are updated by building a covisibility graph, which can be optimized 
using tools like g2o~\cite{Kuemmerle11icra} or gtsam~\cite{Dellaert12tr}. We use the initial poses and sparse map points as inputs to the back-end of the system. Typically, the visual odometry tracks a 
image frame within 20 to 30 milliseconds. Whenever a keyframe is detected, it is added to the Bundle Adjustment (BA) of the local map. During this period, the back-end of the system continues to optimize the current 3D Gaussian map. The tracking module of the front-end usually extracts enough keypoints from texture-rich areas, which are often characterized of local concentration and global sparseness. While local concentration is crucial for reconstructing detailed scenes, relying solely on these keypoints make it difficult to achieve high-quality global reconstruction. Furthermore, to meet the real-time requirements, the number of keypoints and keyframes is limited.

\subsection{{Patch-Grid Sampling}}
\label{subsec:Patch-Grid}
\begin{figure}[t]
	\centering
    \subfigure[Active Keypoints\label{fig:active_keypoints}]{\includegraphics[width=0.23\textwidth]{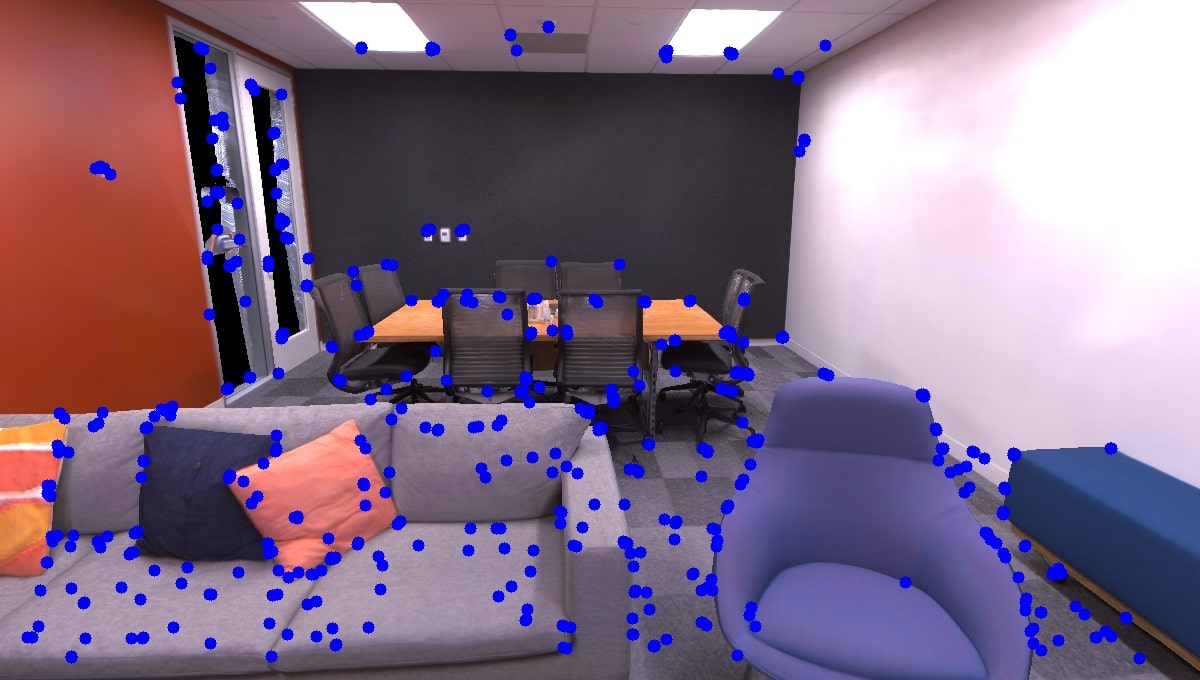}}\hspace{5pt}
	\subfigure[Inactive Keypoints\label{fig:inactive_keypoints}]{\includegraphics[width=0.23\textwidth]{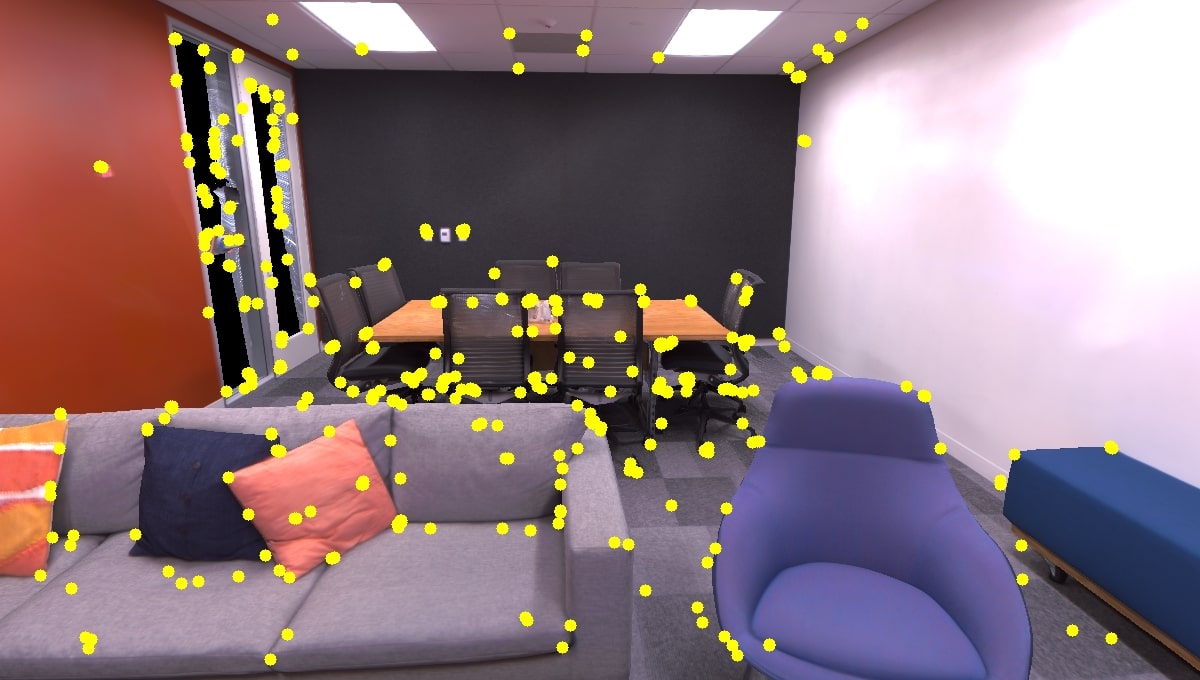}}\\
	\subfigure[Sample Points\label{fig:sample_points}]{\includegraphics[width=0.23\textwidth]{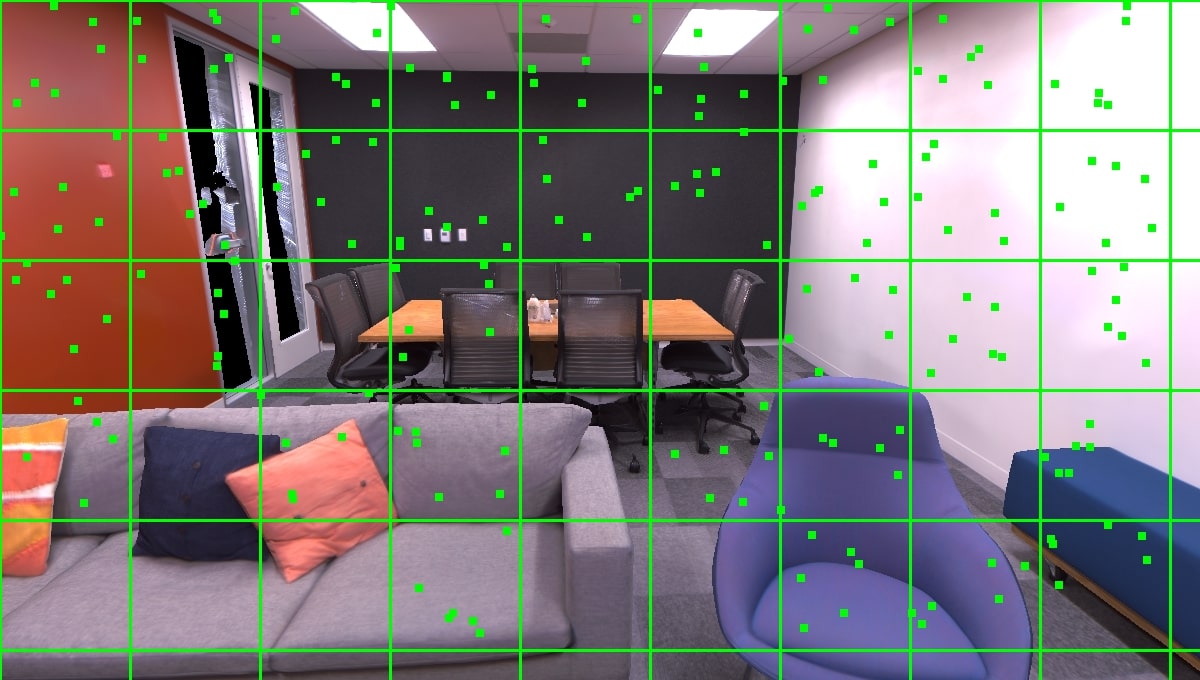}}\hspace{5pt}
	\subfigure[Initial Points\label{fig:initial_points}]{\includegraphics[width=0.23\textwidth]{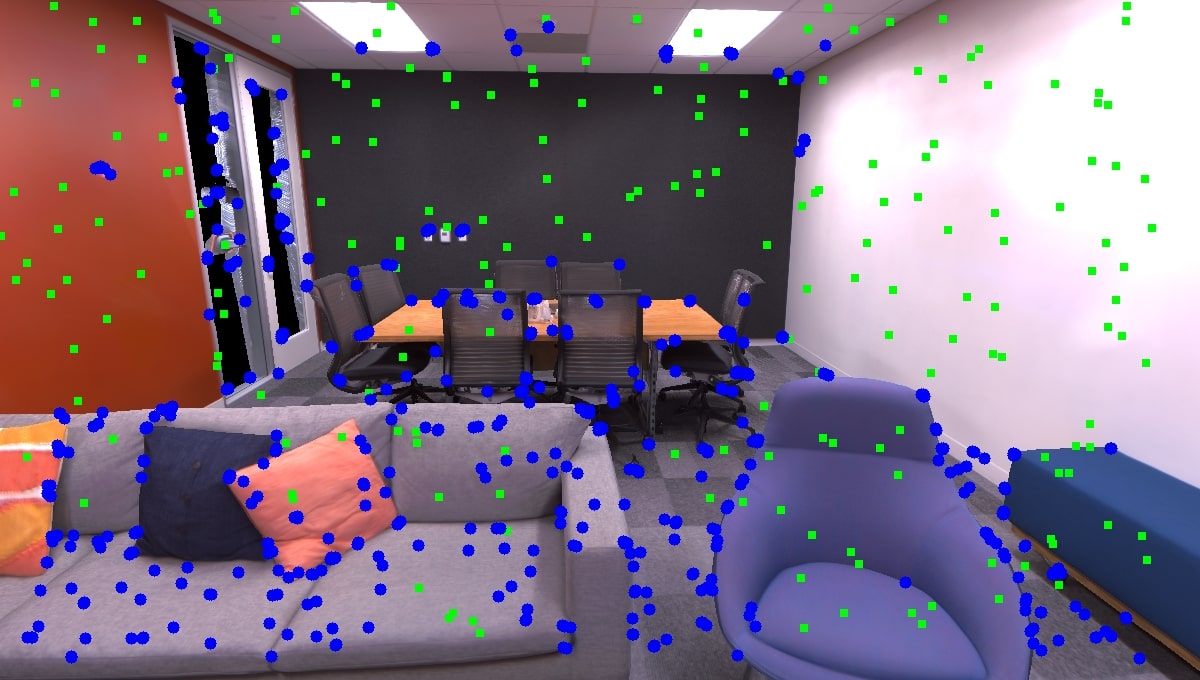}}
	\caption{Initial points from RGB image and used for 3DGS~\cite{kerbl3Dgaussians}.} 
    \label{fig:initial points}
    \vspace{-3.0em}
\end{figure}

Generally, more 3D Gaussian primitives can represent the scene more accurately and with greater complexity, resulting in higher rendering quality. In Sec.~\ref{subsec:Tracking} as the front-end to detect keyframes and extract keypoints. As introduced in~\cite{hhuang2024photoslam}, fewer than 30\% of the feature points in an image frame are active (Fig.~\ref{fig:active_keypoints}): they have corresponding 3D points. Most of them are inactive (Fig.~\ref{fig:inactive_keypoints}). But they are concentrated in areas with rich textures. 
Even through we perform subsequent cloning or splitting of 3D Gaussian primitives with large loss gradients, the keypoints remain sparse and may not sufficiently model the entire scene. To address this issue, we introduce Patch-Grid (PG) sampling to densify the 3D Gaussian primitives. 
The PG is a 2D grid that divides the image into patches, each containing a set of keypoints or sampling points. Only when the number of keypoints is below a certain threshold, we use the PG to uniformly sample points in these patches (Fig.~\ref{fig:sample_points}). When its depth is not available, use the depth of the nearest neighbor keypoints, such as monocular. All points are subsequently used to initialize the 3D Gaussian primitives.

\subsection{{3D Gaussian Splatting}}
\label{subsec:GS Splatting}
The 3D Gaussian map ${G}$ contains a large number of Gaussians ${G_i}$: $G=\left\{G_i\left(\mu_i, \Sigma_i, o_i, c_i\right) \mid i=1, \ldots, N\right\}$, where $\mu_i \in {\mathbb{R}^{3}}$ and $\Sigma_i \in {\mathbb{R}^{9}}$ are the mean and covariance, $o_i \in \mathbb{R}$ is the opacity,  and color $c_i$ can be converted from spherical harmonics ${SH \in \mathbb{R}^{16}}$. The 3D Gaussian map is rendered into image by projecting the 3D Gaussian primitives from world space into the image space throuth a projection transformation:
\begin{equation}
  {\mu}_I=\pi\left({T}_{CW} \cdot {\mu}_W\right), {\Sigma}_I={J} {W} {\Sigma}_W {W}^T {J}^T,
  \label{eq:projection_matrix}
\end{equation}
where $\mathcal{N}(\mu_I, \Sigma_I)$ and $\mathcal{N}(\mu_W$, $\Sigma_W)$ are the mean and covariance of the Gaussian in the image and world space, respectively. $\pi\left(\cdot\right)$ is the projection function, $T_{CW}$ is the camera pose, $J$ is the Jacobian matrix of the linear approximation of the projection transformation, and $W$ is the rotation part of $T_{CW}$.
As for the covariance $\Sigma_W$, which can be parameterized by the scale $S \in {\mathbb{R}^{3}}$ and the rotation matrix $R \in {\mathbb{R}^{9}}$ as:
\begin{equation}
  \Sigma_W=R S S^T R^T .
  \label{eq:cov3D_matrix}
\end{equation}
Notably, throughout the remainder of this paper, we use the covariance matrix in the world coordinate system and no longer use subscripts to denote it.
\subsection{{Voxel-based Merging}}
\label{subsec:voxel_based_merge}
In contrast to~\cite{kerbl3Dgaussians}, which selects 3D Gaussian primitives with large loss gradients for cloning or splitting, we merge 3D Gaussian primitives with small loss gradients in the current local map. Then the mask ${M(G)}$ is defined as:  
\begin{equation}
  M(G)=\left[\nabla G<\tau \wedge K_{\min } \leq K \leq K_{\max }\right],
  \label{eq:grad_mask}
\end{equation}
where $\nabla{G}$ represents the average gradient of each Gaussian primitive $G_i$ in 3D space, and the gradient threshold ${\tau = 0.001 }$ is used to identify primitives with more stable geometry and appearance. The hyperparameters ${K_{min}}$ and ${K_{max}}$ denote the start and end keyframe indices in the current keyframe list ${K}$. After that, we select the 3D Gaussian primitives that need to merge in the voxel space. 

Before that, we need to determine the number of 3D Gaussian primitives in each voxel. Since we have narrowed the search range using the mask in Eq.~\ref{eq:grad_mask}, this process can be completed quickly within about 3 milliseconds. The number of 3D Gaussian primitives $G_i$ in a voxel is affected by the voxel size. 
In general, if there are ${(2, 3, 4\dots)}$ 3D Gaussian primitives in a voxel, there will be ${(1, 3, 6\dots)}$ matching pairs for which we need to calculate the Mahalanobis Distance (MD).  

\textbf{Merge 3D Gaussians Position.}
Since the similarity between two Gaussians increases as their Mahalanobis Distance (MD) decreases, we first compute the squared MD between the 3D Gaussians $\mathcal{N}(\mu_i, \Sigma_i)$ and $\mathcal{N}(\mu_j, \Sigma_j)$ as follows:
\begin{equation}
  d_M(\mu_j ; \mathcal{N}(\mu_i, \Sigma_i))^2=(\mu_j - \mu_i)^T \Sigma_i^{-1}(\mu_j - \mu_i).
  \label{eq:calculate_MD}
\end{equation}
In this context, ${i}$ and ${j}$ represent the indices at which 3D Gaussian primitives are inserted into the voxel. More precisely, ${G_j}$ denotes a recently inserted primitive, whereas ${G_i}$ refers to one that was already present in the voxel. 
Then following Eq.~\ref{eq:calculate_MD}, we calculate the minimum ${d_M}$ for the 3D Gaussian pair ${\{G_i, G_j\}}$ within the voxel. We implement this in custom CUDA kernels to accelerate per-voxel computations to milliseconds.

Subsequently, we refer to the chi-square distribution table to select the critical value ${\chi_{0.05}^2 = 7.815}$, which corresponds to 3 degrees of freedom and a $95\%$ confidence level. The two Gaussians are considered similar only if the MD is bellow this critical value. The optimal mean $\boldsymbol{\mu_k^{\ast}}$ is computed as:
\begin{equation}
  \boldsymbol{\mu_k^{\ast}}=\mathop{\arg\max}\limits_{\boldsymbol{\mu_k}}\left(\log f\left(\boldsymbol{\mu_k} ; {\Sigma_i}\right)+\log f\left(\boldsymbol{\mu_k}; {\Sigma_j}\right)\right),
  \label{eq:optimal_mean}
\end{equation}
where the function ${f(\mathord{\cdot})}$ denotes the Probability Density Function (PDF) of the Gaussian distribution.

By constructing a maximized joint log-likelihood function to find intersection of two independent Gaussian distributions within the same voxel (see Eq.~\ref{eq:optimal_mean}), we then obtain an analytical solution:
\begin{equation}
  \boldsymbol{\mu_k^{\ast}} =\left(\Sigma_i^{-1}+\Sigma_j^{-1}\right)^{-1}\left(\Sigma_i^{-1} \mu_i+\Sigma_j^{-1} \mu_j\right).
  \label{eq:optimal_mean_val}
\end{equation}
\textbf{Merge 3D Gaussians Scale and Rotation.}
For the newly merged 3D Gaussian primitive ${G_k}$, its spatial shape is determined by the Wasserstein-2 distance~\cite{givens1984class}, 
which measures how different the probability distribution of ${G_k}$ is from those of ${G_i}$ or ${G_j}$,
\textit{e.g.}, the Wasserstein-2 distance between two Gaussians ${\{G_k, G_i\}}$ is defined as:
\begin{equation}
  W_2(\mathcal{N}\left(\mu_k, \Sigma_k\right) ; \mathcal{N}\left(\mu_i, \Sigma_i\right))^2:=\inf \mathbb{E}\left(\|G_k-G_i\|_2^2\right).
  \label{eq:Wasserstein-2}
\end{equation}
Next, assume that the centers ${\{u_i, u_j\}}$ of two Gaussians ${\{G_i, G_j\}}$ are close to each other within the same voxel, or even ideally share the same position ${u_k}$ in voxel space. In this case, we only need to consider their uncertainty, namely the covariance matrix ${\{\Sigma_i, \Sigma_j\}}$. Consequently, Eq.~\ref{eq:Wasserstein-2} can be reformulated as: ${d:=W_2\left(\mathcal{N}\left(0, \Sigma_k\right) ; \mathcal{N}\left(0, \Sigma_i\right)\right)}$. Following the derivation detailed by Givens \textit{et al.}~\cite{givens1984class}, this expression for two Gaussians ${\{G_k, G_i\}}$ can be simplified to:
\begin{equation}
  d_{W_2}(\Sigma_k; \Sigma_i)^2=\operatorname{tr}\left(\Sigma_k+\Sigma_i-2\left(\Sigma_k^{1 / 2} \Sigma_i \Sigma_k^{1 / 2}\right)^{1 / 2}\right).
\end{equation}
In this work, we use the limited-memory Broyden-Fletcher-Goldfarb-Shanno (L-BFGS) algorithm to optimize the covariance matrix ${\Sigma_k}$ of the merged Gaussian ${G_k}$ by solving the following minimization problem:
\begin{equation}
  \boldsymbol{{\Sigma_k}^{\ast}}=\mathop{\arg\min}\limits_{\boldsymbol{\Sigma_k}}\left(d_{W_2}\left(\boldsymbol{\Sigma_k}; \Sigma_i\right)^2+d_{W_2}\left(\boldsymbol{\Sigma_k}; \Sigma_j\right)^2\right).
\end{equation}
The L-BFGS algorithm is particularly well-suited for this task due to its ability to handle the smooth, continuous nature of the Wasserstein-2 distance. This approach ensures convergence to a satisfactory solution while balancing efficiency and accuracy.
For the rotation matrix ${R}$ and scale ${S}$ in Eq.~\ref{eq:cov3D_matrix}, we represent the rotation using a quaternion $\boldsymbol{q}$ and the scaling using a vector $\boldsymbol{s}$, following~\cite{kerbl3Dgaussians}. 

In order to further improve the solver speed and find the optimal rotation ${q_k^{\ast}}$ and scale ${s_k^{\ast}}$, we initialize their values at the start of each iteration. The initial rotation $\boldsymbol{q_{k, 0}}$ is computed using the spherical linear interpolation (slerp) between ${q_i}$ and ${q_j}$, while the initial scale $\boldsymbol{s_{k, 0}}$ is set to the average of ${s_i}$ and ${s_j}$. These are calculated as follows:
\begin{equation}
  \begin{cases} 
    \boldsymbol{{q}_{k, 0}} = \boldsymbol{slerp}\left({q_i, q_j}, t\right) \\
    \boldsymbol{{s}_{k,0}} = \left({s_i + s_j}\right) /2 + \left[\delta_1, \delta_2, \delta_3\right]^T.
  \end{cases}
  \label{eq:initial_rotation}
\end{equation}
The hyperparameter ${t}$ is used to control the interpolation between two quaternions ${\{q_i, q_j\}}$, and its range is ${(0, 1)}$. We set ${t = 0.5}$, in our experiments. Additionally, to mitigate numerical instability during eigenvalue decomposition (EVD), we introduce small offsets ${\delta_1 = 0.001}$, ${\delta_2 = 0.002}$, ${\delta_3 = 0.003}$. The process is then performed iteratively until the convergence criterion is met. Since the initial values are already provided by Eq.~\ref{eq:initial_rotation}, the solver converges more quickly after just a few iterations. The entire process can be completed on custom CUDA kernels.

\textbf{Merge 3D Gaussians color and opacity.}
In 3DGS~\cite{kerbl3Dgaussians}, densification is performed by cloning or splitting, and these new 3D Gaussian primitives initially inherit the same color and opacity as the original ones. Similarly, in our approach, the new 3D Gaussian primitives ${G_k}$ resulting from merging preserve the color and opacity of the older 3D Gaussian ${G_i}$, which generally corresponds to the one that was cloned or split. We then continue to optimize them in subsequent steps as described in Sec.~\ref{subsec:Mapping}.

As shown in the left part of Fig.~\ref{fig:system flowchart}, we use red dashed boxes to mark two pairs of geometrically similar 3D Gaussian primitives and identify the corresponding ${G_k}$ (indicated by the red arrows) via the process described above. To enhance distinction, we assign them different colors; however, their actual colors are nearly the same within the same voxel, as illustrated in the right part of Fig.~\ref{fig:system flowchart}.

\subsection{{Mapping}}
\label{subsec:Mapping}
In the back-end of our proposed system, we focus on the rendering process after the splatting step introduced in Sec.~\ref{subsec:GS Splatting} and in Sec.~\ref{subsec:voxel_based_merge}, which is responsible for updating the 3D Gaussian primitives in voxel space. The 3D Gaussian primitives can be rasterized into the image ${I_\mathrm{r}}$ using tile-based rendering. The pixel color ${C_p}$ and depth ${D_p}$ are then synthesized by blending ${N}$ Gaussians with the keyframe poses ${T_{C W}}$, as follows:
\begin{equation}
  \begin{aligned}
    C_{p} = \sum_{i \in N} {c}_i \alpha_i \prod_{j=1}^{i-1}\left(1-\alpha_i\right),
    D_{p} = \sum_{i \in N} {d}_i \alpha_i \prod_{j=1}^{i-1}\left(1-\alpha_i\right),
  \end{aligned}
  \label{eq:rendering_and_depth}
\end{equation}
where ${\alpha_i}$ is defined as $\sigma_i \cdot {G}\left(T_{CW}, \mathbf{\mu}_i, \mathbf{q}_i, \mathbf{s}_i\right)$, and ${\sigma_i}$ represents the density, as used in~\cite{nerf} and~\cite{kerbl3Dgaussians}.
The depth ${d_i}$ of the 3D Gaussian primitive ${G_i}$ in the camera coordinate system can be compared with the ground truth depth image to calculate the geometric residual. 

In order to encourage sphericity and avoid the issue of 3D Gaussian primitives becoming highly elongated along the viewing direction, we introduce the isotropic Gaussian loss term ${\mathcal{L}_{\text {iso}}}$, following~\cite{Matsuki_2024_CVPR}.
Finally, the optimization process is performed by minimizing the loss function ${\mathcal{L}}$:
\begin{equation}
  \mathcal{L}=(1-\lambda)\left|I_{\mathrm{r}}-I_{\mathrm{gt}}\right|_1+\lambda\left(1-\operatorname{SSIM}\left(I_{\mathrm{r}}, I_{\mathrm{gt}}\right)\right) + \mathcal{L}_{\text {iso}},
  \label{eq:loss_function}
\end{equation}
where $\lambda$ is the weight of the structural similarity (SSIM) loss, and $I_{\mathrm{gt}}$ denotes the ground truth RGB image.
Additionally, if the depth is available (RGB-D), we can also render per-pixel depth by alpha-blending, as expressed in Eq.~\ref{eq:rendering_and_depth},  and add the geometric residual, as defined in Eq.~\ref{eq:loss_function}.       

\section{{Experiments}}
\label{sec:Experiments}
In this section, we first introduce the datasets used and provide the implementation details of our approach (Sec.~\ref{subsec:Evaluation Setup}). We then compare our system with other state-of-the-art (SOTA) 3DGS-based SLAM methods in various scenarios encapsulating monocular and RGB-D (Sec.~\ref{subsec:Results and Evaluation}). Finally, we deploy our system on an embedded device to demonstrate its real-time performance (Sec.~\ref{subsec:Real-World Experiments}).

\begin{figure}[t]
	\centering
	\subfigure[Ground Truth]{\includegraphics[width=0.23\textwidth]{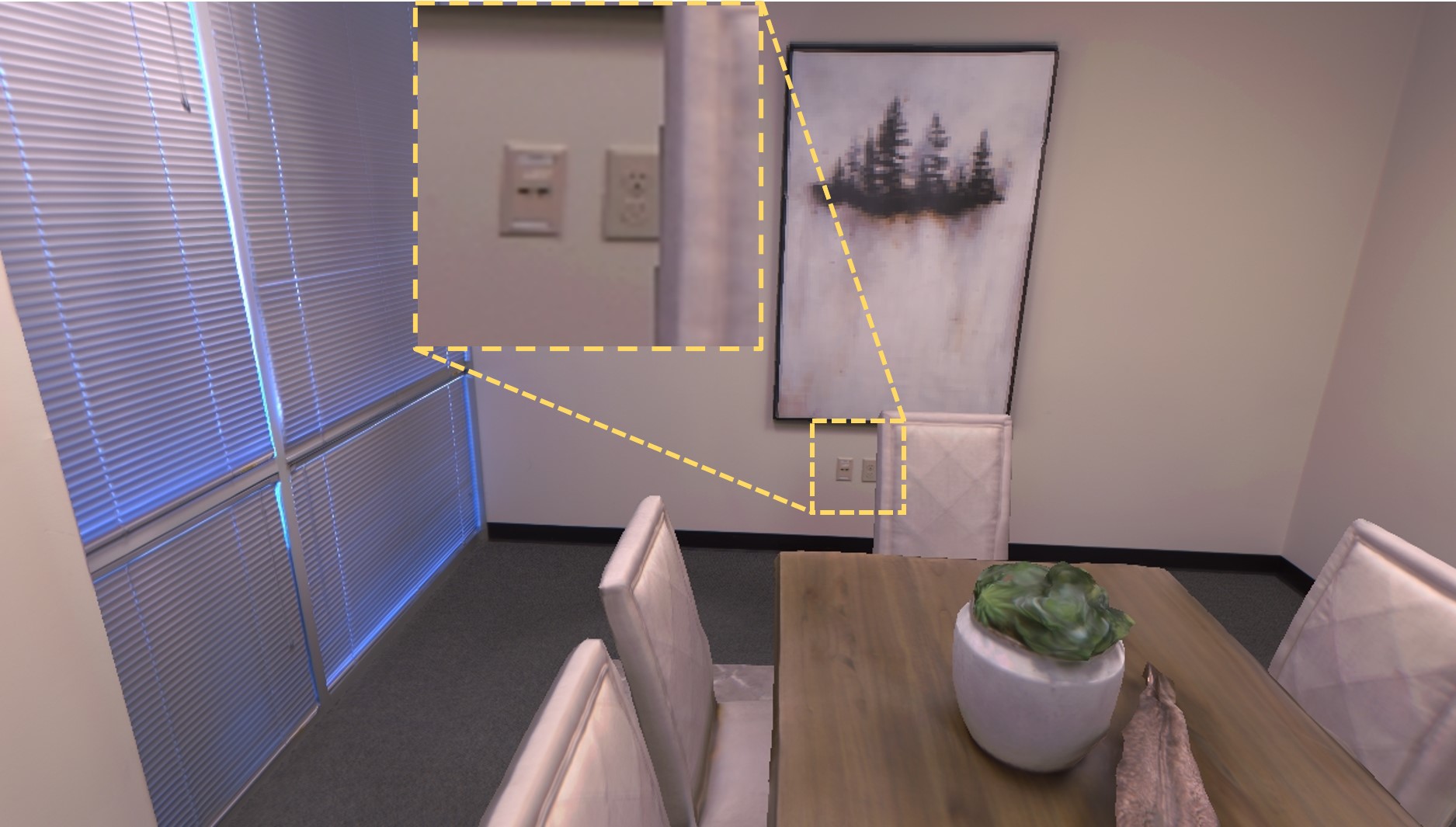}}
	\hspace{5pt}
	\subfigure[MonoGS~\cite{Matsuki_2024_CVPR}]{\includegraphics[width=0.23\textwidth]{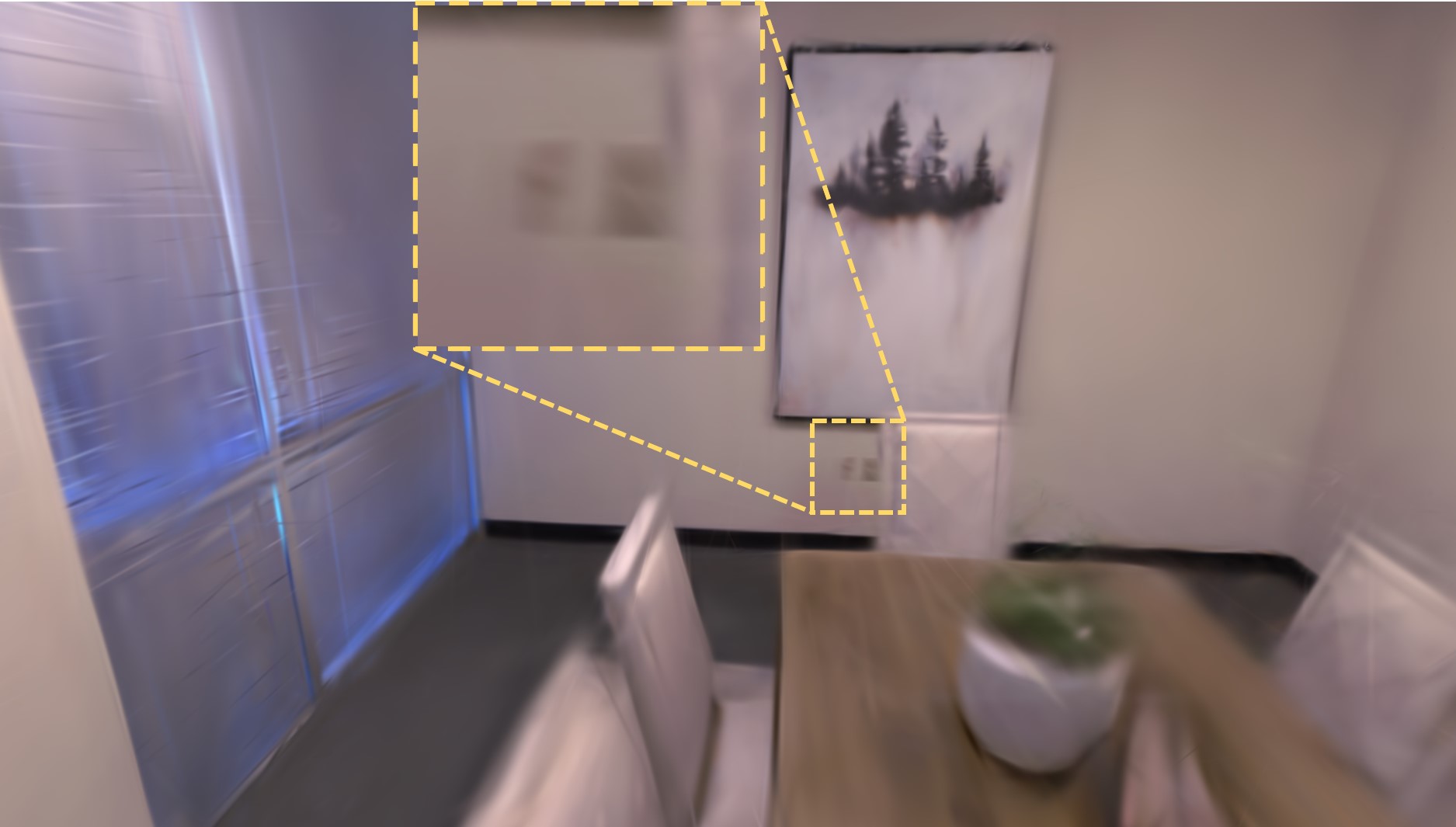}}
	\subfigure[Photo-SLAM~\cite{hhuang2024photoslam}]{\includegraphics[width=0.23\textwidth]{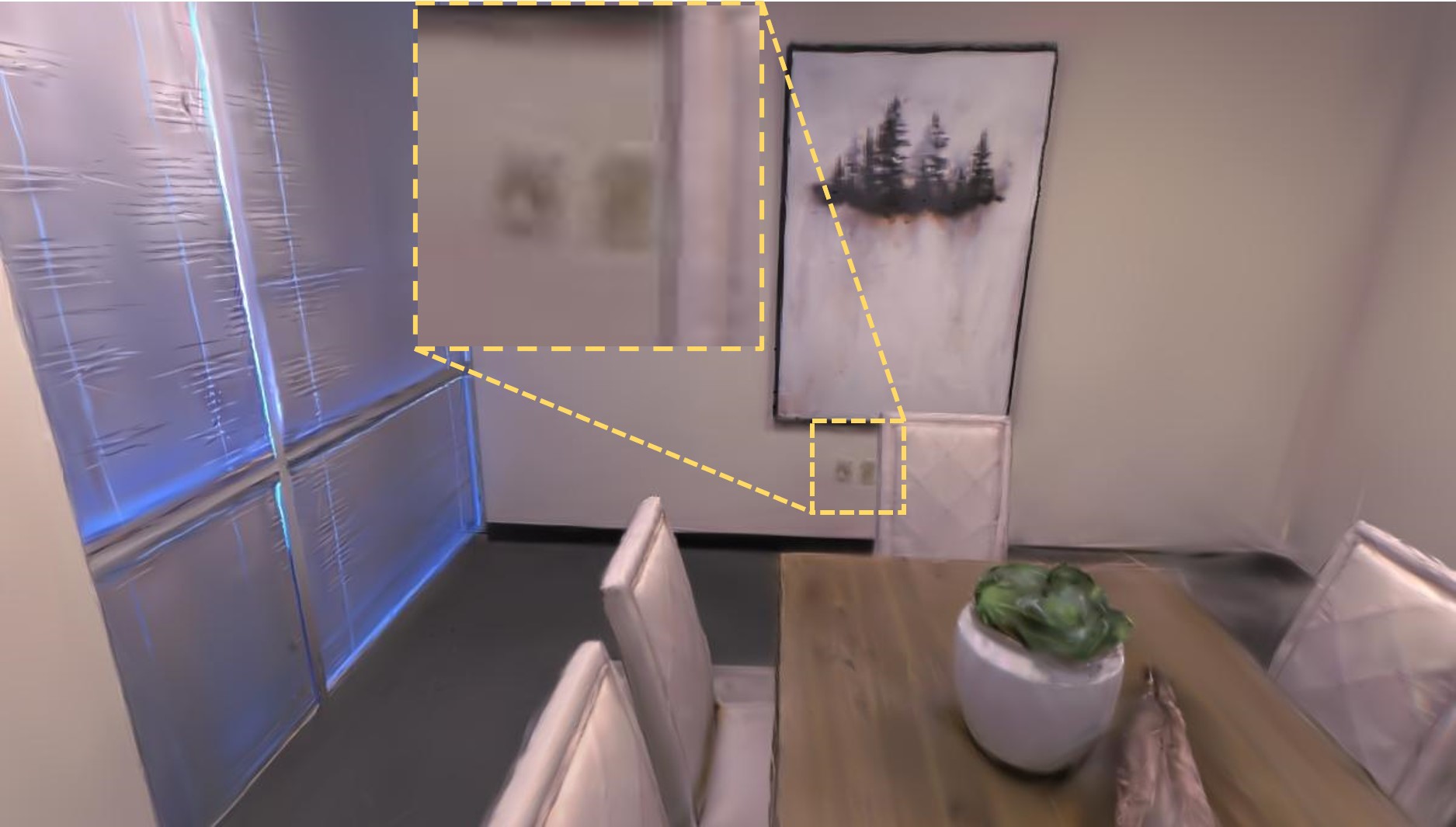}}
	\hspace{5pt}
	\subfigure[\textbf{Ours}]{\includegraphics[width=0.23\textwidth]{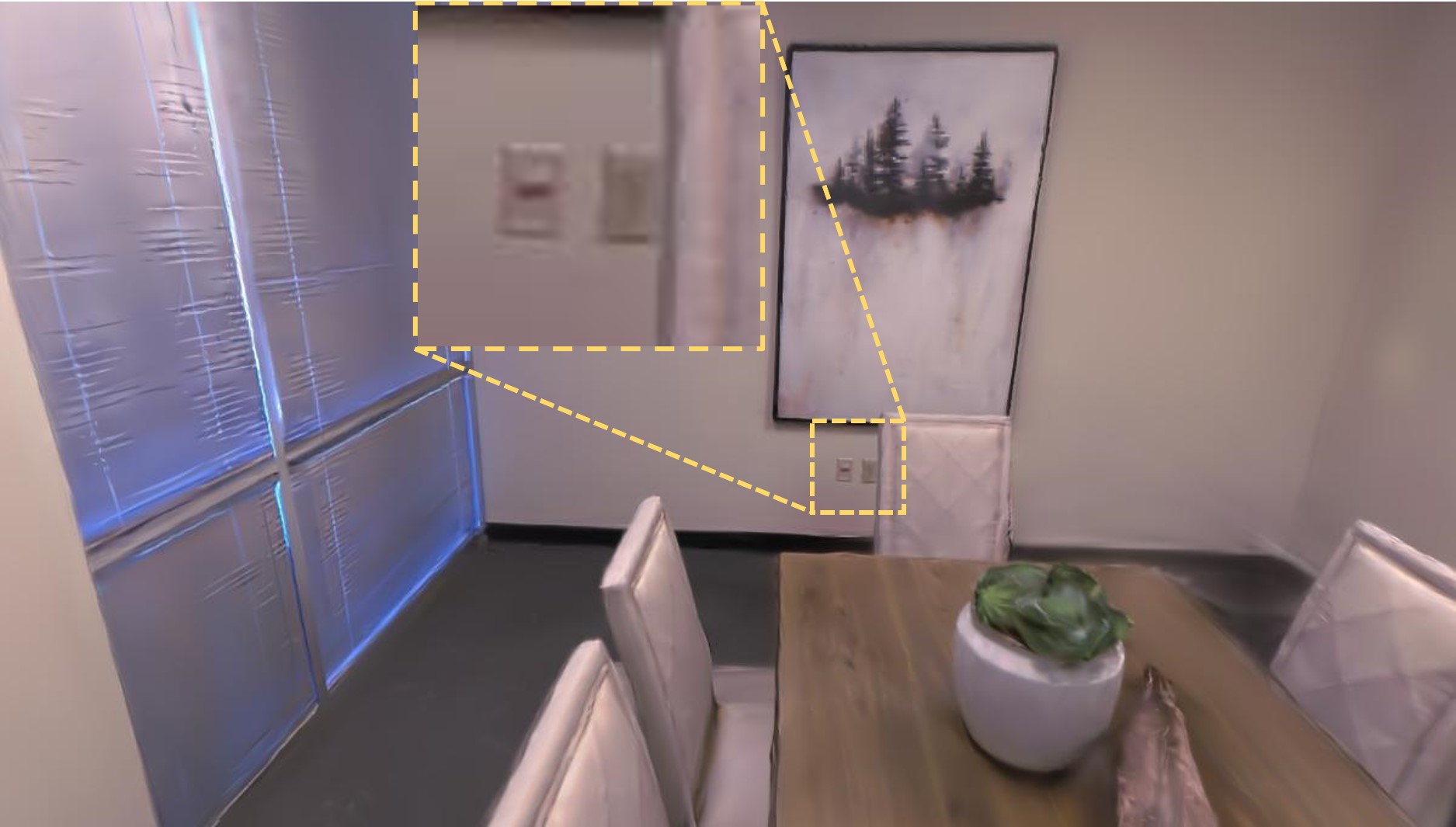}}
	\caption{Qualitative comparison on monocular of Replica.} 
    \label{fig:replica_monocular}
    \vspace{-3.0em}
\end{figure}

\subsection{{Evaluation Setup}}
\label{subsec:Evaluation Setup}
\textbf{Datasets.}
We perform quantitative evaluations on both Replica~\cite{dataset2019replica} and TUM RGB-D~\cite{dataset2012tum} datasets, following~\cite{Keetha_2024_CVPR, Matsuki_2024_CVPR, ha2024rgbdgsicpslam, hhuang2024photoslam}. For real-world experiments, we use an Intel RealSense D435i RGB-D camera to collect an indoor dataset in real-time, as shown in Fig.~\ref{fig:eye_catcher}.

\textbf{Baselines.}
First, Photo-SLAM~\cite{hhuang2024photoslam} uses the same front-end~\cite{campos2021orb3} for visual odometry as our approach. Second, SplaTAM~\cite{Keetha_2024_CVPR} is the first open-source framework. Third, MonoGS~\cite{Matsuki_2024_CVPR} is notable for being the first to support monocular cameras. Finally, recently developed 3DGS-based real-time SOTA methods include GS-ICP SLAM~\cite{ha2024rgbdgsicpslam}.

\textbf{Evaluation Metrics.}
We use FPS to evaluate the real-time performance of the system. For rendering quality, we use PSNR (in dB) to measure the quality of rendered images. As for system resources, we measure the amount of GPU-allocated memory (in GB) once the SLAM system has finished.
At the same time, we also recorded the number of points in the final saved PLY file (in thousands).
To mitigate the effects of different system configurations, we run each sequence multiple times (3 groups of valid values) and report the average results. In all tables, we highlight the top two results in bold and underline. The arrow indicates whether higher or lower values are better for each metric.

\begin{table}[t]
    \centering
    \footnotesize
    \tabcolsep=0.1cm 
    {
    \begin{tabular}{cc|ccccc}
    \toprule
    \multicolumn{2}{c}{\textbf{On Replica Dataset}}& \multicolumn{4}{c}{\textbf{office (0 \textasciitilde 4) Avg.}} \\
    \cmidrule(lr){1-2}\cmidrule(lr){3-6}
	\textbf{Cam}&\textbf{Method} & \textbf{FPS${\uparrow}$} & \textbf{PSNR}${\uparrow}$ & \textbf{Mem.}${\downarrow}$ & \textbf{Points}${\downarrow}$ \\
	\midrule      
    \multirow{3}{*}{{\rotatebox{90}{Mono}}}
    &MonoGS~\cite{Matsuki_2024_CVPR}  &1.526  &27.968 &10.642 &108.226  \\
    &Photo-SLAM~\cite{hhuang2024photoslam} &\textgreater30  &\underline{33.585} &\underline{3.142}  &\underline{81.869}\\
    &\textbf{Ours}&\textgreater30  &\bf{33.955} &\bf{2.748} &\bf{80.163} \\
    \midrule\midrule
    \multirow{5}{*}{{\rotatebox{90}{RGB-D}}}
    &SplaTAM~\cite{Keetha_2024_CVPR} &{0.156}  &34.264 &8.002 &5408.706 \\
    &MonoGS~\cite{Matsuki_2024_CVPR} &1.039  &\underline{37.287} &12.489 &302.739 \\
    &GS-ICP SLAM \cite{ha2024rgbdgsicpslam} &\textgreater 30 &\bf{39.043} &{4.105} &{1644.054} \\
    &Photo-SLAM~\cite{hhuang2024photoslam} &\textgreater 30  &36.084  &\underline{2.518} &\underline{100.534}\\
    &\textbf{Ours} &\textgreater 30  &{37.150} &\bf{1.952} &\bf{98.214} \\
    \bottomrule
    \end{tabular}}
    \caption{Quantitative results on the Replica offices.} 
    \label{tab:replica_eval_office}
    \vspace{-0.0em}
\end{table}
\textbf{Implementation Details.}
We implemented our framework fully in C++ and CUDA, and then evaluated our system on a desktop computer equipped with an Intel Core i9-14900K CPU and a single NVIDIA RTX 4080 SUPER 16GB GPU. All baselines were tested on our platform using their official code on public datasets. Furthermore, we conduted additional experiments on a Jetson AGX Orin 64GB Developer Kit.

\begin{table}[t]
    \centering
    \footnotesize
    \tabcolsep=0.1cm 
    {
    \begin{tabular}{cc|ccccc}
    \toprule
    \multicolumn{2}{c}{\textbf{On Replica Dataset}}& \multicolumn{4}{c}{\textbf{room (0 \textasciitilde 2) Avg.}} \\
    \cmidrule(lr){1-2}\cmidrule(lr){3-6}
	\textbf{Cam}&\textbf{Method} & \textbf{FPS${\uparrow}$} & \textbf{PSNR}${\uparrow}$ & \textbf{Mem.}${\downarrow}$ & \textbf{Points}${\downarrow}$ \\
	\midrule      
    \multirow{3}{*}{{\rotatebox{90}{Mono}}}
    &MonoGS~\cite{Matsuki_2024_CVPR}  &1.368  &24.712 &11.542 &119.242  \\
    &Photo-SLAM~\cite{hhuang2024photoslam} &\textgreater30  &\underline{27.541} &\underline{2.560}  &\underline{72.288}\\
    &\textbf{Ours}&\textgreater30  &\bf{28.852} &\bf{1.972} &\bf{71.458} \\
    \midrule\midrule
    \multirow{5}{*}{{\rotatebox{90}{RGB-D}}}
    &SplaTAM~\cite{Keetha_2024_CVPR} &{0.142}  &33.978 &7.706 &5844.432 \\
    &MonoGS~\cite{Matsuki_2024_CVPR} &0.947  &\underline{35.180} &13.509 &328.565 \\
    &GS-ICP SLAM \cite{ha2024rgbdgsicpslam} &\textgreater 30 &\bf{36.509} &4.098 &1626.239 \\
    &Photo-SLAM~\cite{hhuang2024photoslam} &\textgreater30  &30.486  &\underline{2.439} &\underline{113.253}\\
    &\textbf{Ours} &\textgreater30  &31.013 &\bf{1.957} &\bf{110.362} \\
    \bottomrule
    \end{tabular}}
    \caption{Quantitative results on the Replica rooms.} 
    \label{tab:replica_eval_room}
    \vspace{-8.0em}
\end{table}
\subsection{{Results and Evaluation}}
\label{subsec:Results and Evaluation}
\textbf{Quantitative and Qualitative Results.}
We present the quantitative results in tables, while the figures show the qualitative results.
On Replica datasets, we use 8 sequences: office (0 \textasciitilde 4) and room (0 \textasciitilde 2). The results in Tab.~\ref{tab:replica_eval_office} and Tab.~\ref{tab:replica_eval_room} report the average performance for 5 office sequences and 3 room sequences, respectively. On TUM RGB-D dataset, we show detailed results for three sequences in Tab.~\ref{tab:tum_eval}: fr1-desk, fr2-xyz and fr3-office.

\begin{table*}[t]
    \centering
    \footnotesize
    \tabcolsep=0.08cm 
    {
    \begin{tabular}{cc|cccccccccccc}
    \toprule
	\multicolumn{2}{c}{\textbf{On TUM Dataset}}&  \multicolumn{4}{c}{\textbf{fr1-desk}}& \multicolumn{4}{c}{\textbf{fr2-xyz}}& \multicolumn{4}{c}{\textbf{fr3-office}}\\\cmidrule(lr){1-2} \cmidrule(lr){3-6} \cmidrule(lr){7-10}\cmidrule(lr){11-14}
	{\scriptsize \textbf{Cam}}&{\scriptsize \textbf{Method}}&{\scriptsize \textbf{FPS} $\uparrow$} &{\scriptsize \textbf{PSNR} $\uparrow$} &{\scriptsize \textbf{Mem.} $\downarrow$} & {\scriptsize \textbf{Points} $\downarrow$}&{\scriptsize \textbf{FPS} $\uparrow$}& {\scriptsize \textbf{PSNR} $\uparrow$} &{\scriptsize \textbf{Mem.} $\downarrow$} & {\scriptsize \textbf{Points} $\downarrow$}&{\scriptsize \textbf{FPS} $\uparrow$}& {\scriptsize \textbf{PSNR} $\uparrow$} &{\scriptsize \textbf{Mem.} $\downarrow$ }& {\scriptsize \textbf{Points} $\downarrow$} \\
	\midrule      
    \multirow{3}{*}{{\rotatebox{90}{Mono}}}
    &MonoGS~\cite{Matsuki_2024_CVPR} &1.914&17.060&2.203&\bf{27.261}&3.224&15.575&\underline{2.802}&\bf{43.646}&2.813&{19.147}&{3.449}&\bf{37.604}\\
    &Photo-SLAM~\cite{hhuang2024photoslam}             
    &\textgreater30&\underline{21.527}&\underline{0.825}&{33.218}&\textgreater30 &\underline{23.407}&{3.123}&{88.923}&\textgreater30&\underline{20.457}&\underline{3.248}&{71.939}\\
    &\textbf{Ours} &\textgreater30&\bf{21.971}&\bf{0.684}&\underline{31.742}&\textgreater30&\bf{23.802}&\bf{2.547}&\underline{85.154}&\textgreater30&\bf{20.851}&\bf{2.762}&\underline{70.243}\\ 
    \midrule\midrule
    \multirow{5}{*}{{\rotatebox{90}{RGB-D}}}
    &SplaTAM~\cite{Keetha_2024_CVPR}  &0.279&\bf{22.892}&{1.331}&{969.957}&0.060&26.288&{9.443}&6323.300&0.297&{21.319}&{3.336}&{806.106}\\
    &MonoGS~\cite{Matsuki_2024_CVPR}  &\underline{1.892}&18.806&2.701&40.586&2.967&15.746&2.429&\bf{31.482}&2.245&19.159&4.070&\bf{53.257}\\ 
    &GS-ICP SLAM~\cite{ha2024rgbdgsicpslam}  &\textgreater30&17.641&1.075&589.856&\textgreater30&{23.108}&\underline{4.168}&{2242.524}&\textgreater30&20.268&3.518&2291.241\\ 
    &Photo-SLAM~\cite{hhuang2024photoslam}  &\textgreater30&21.253&\underline{0.875}&\underline{35.327}&\textgreater30&\underline{26.050}&\underline{0.600}&52.518&\textgreater30&\underline{24.711}&\underline{1.720}&66.638\\ 
    &\textbf{Ours} &\textgreater30&\underline{21.906}&\bf{0.679}&\bf{34.746}&\textgreater30&{\bf{27.421}}&\bf{0.525}&\underline{50.741}&\textgreater30&\bf{24.972}&\bf{1.412}&\underline{64.253}\\
    \bottomrule
    \end{tabular}}
    \caption{Quantitative results on the TUM RGB-D.} 
    \label{tab:tum_eval}
    \vspace{-1.0em}
\end{table*}
\begin{figure*}
    \centering
    \def\imw{0.19}
    \subfigure[SplaTAM~\cite{Keetha_2024_CVPR}]{
    \begin{minipage}{\imw\linewidth}
        \includegraphics[width=\linewidth]{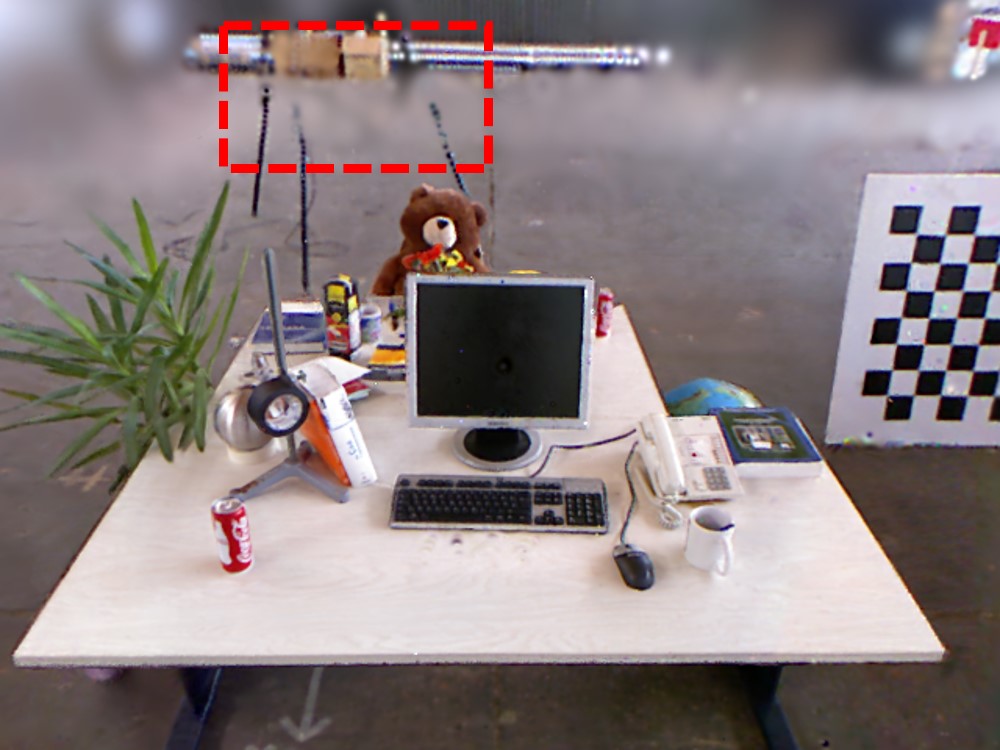}
        \\[0.1cm]
        \includegraphics[width=\linewidth]{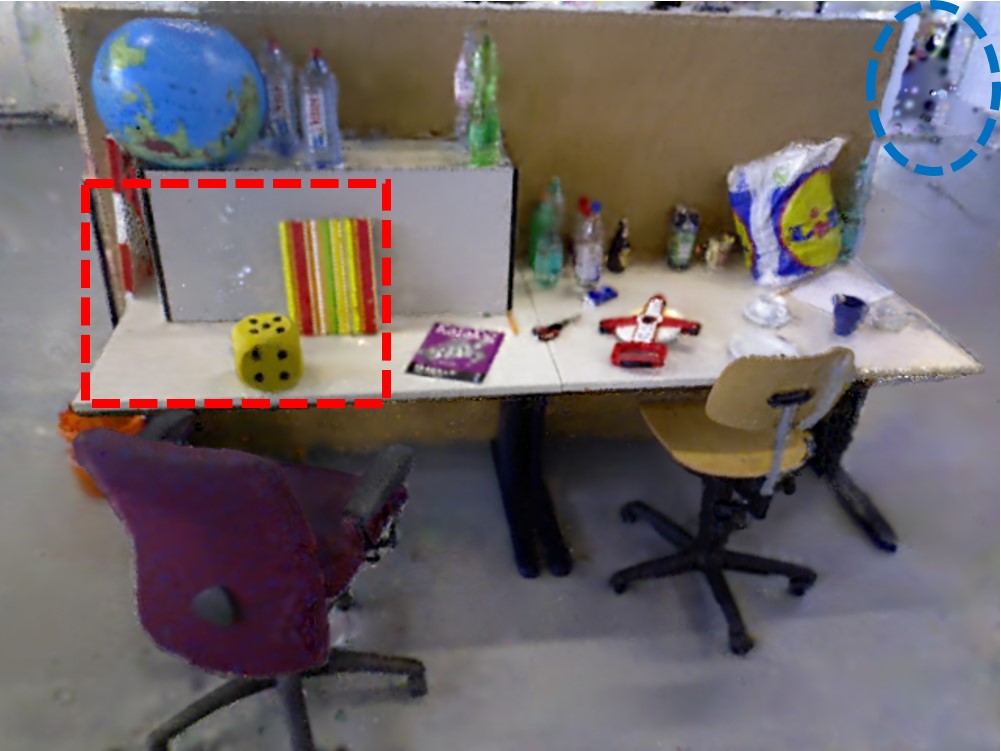}
        \\[-0.2cm]
    \end{minipage}
    }
    \hspace{-0.4cm}
    \subfigure[MonoGS~\cite{Matsuki_2024_CVPR}]{
    \begin{minipage}{\imw\linewidth}
        \includegraphics[width=\linewidth]{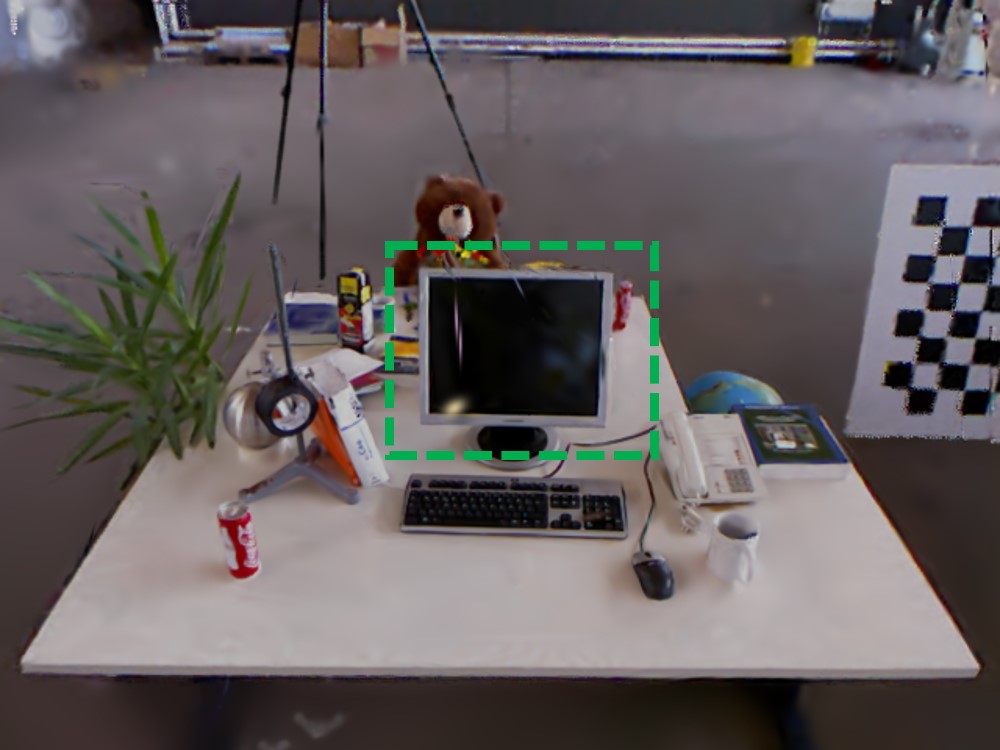}
        \\[0.1cm]
        \includegraphics[width=\linewidth]{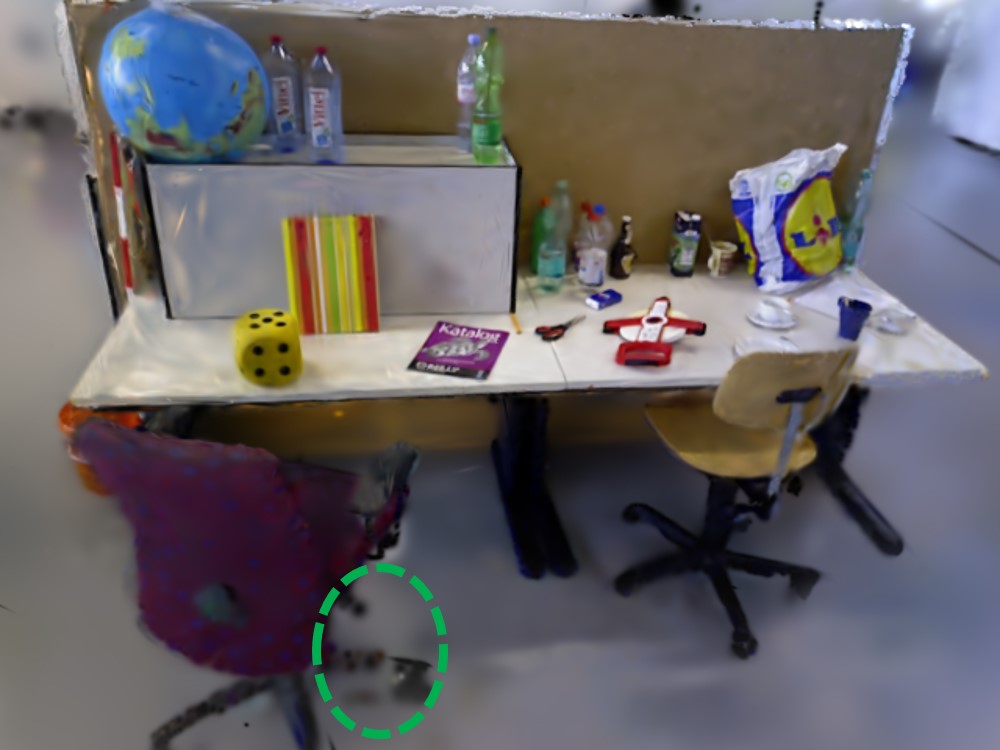}
        \\[-0.2cm]
    \end{minipage}
    }  
    \hspace{-0.4cm}
    \subfigure[GS-ICP~\cite{ha2024rgbdgsicpslam}]{
    \begin{minipage}{\imw\linewidth}
        \includegraphics[width=\linewidth]{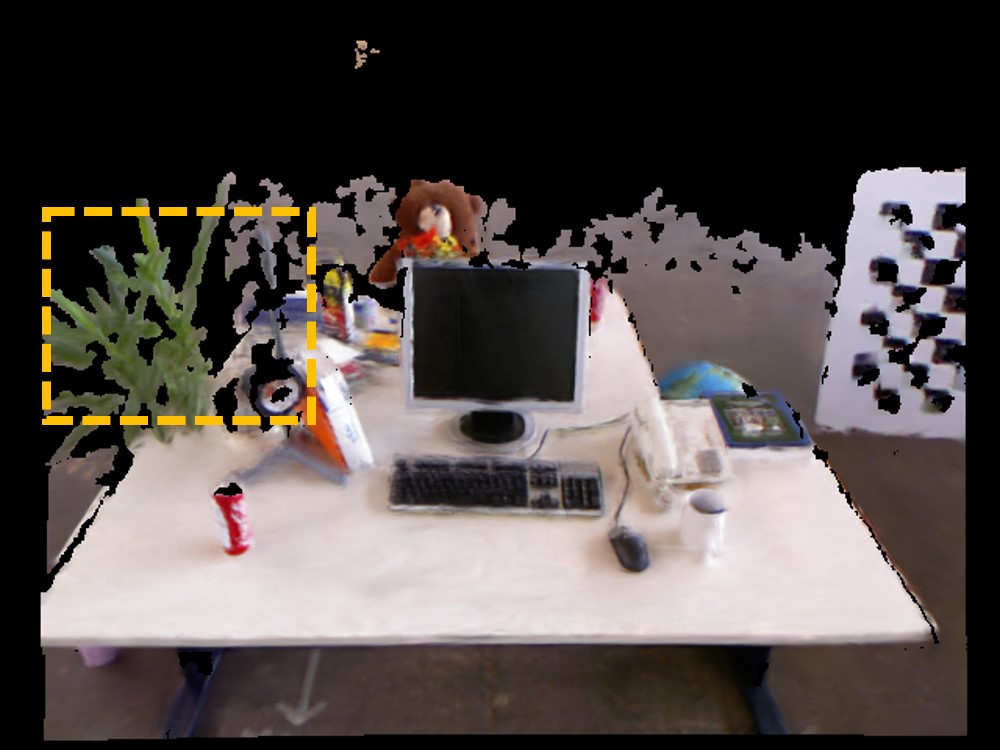}
        \\[0.1cm]
        \includegraphics[width=\linewidth]{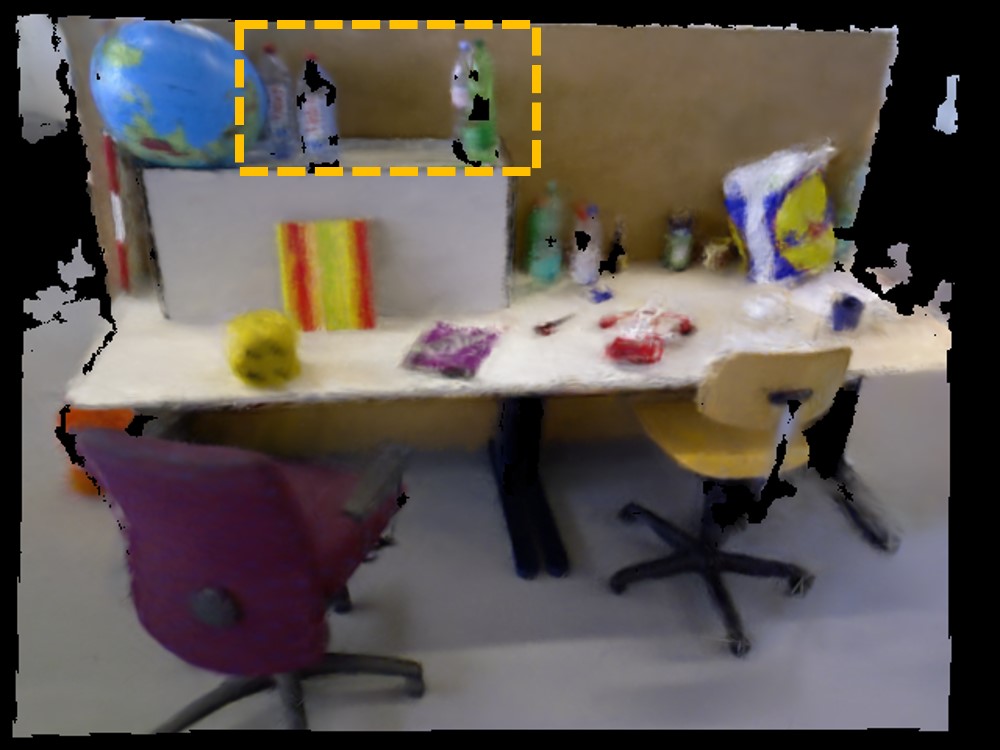}
        \\[-0.2cm]
    \end{minipage}
    }
    \hspace{-0.4cm}
    \subfigure[Photo-SLAM~\cite{hhuang2024photoslam}]{
        \begin{minipage}{\imw\linewidth}
        \includegraphics[width=\linewidth]{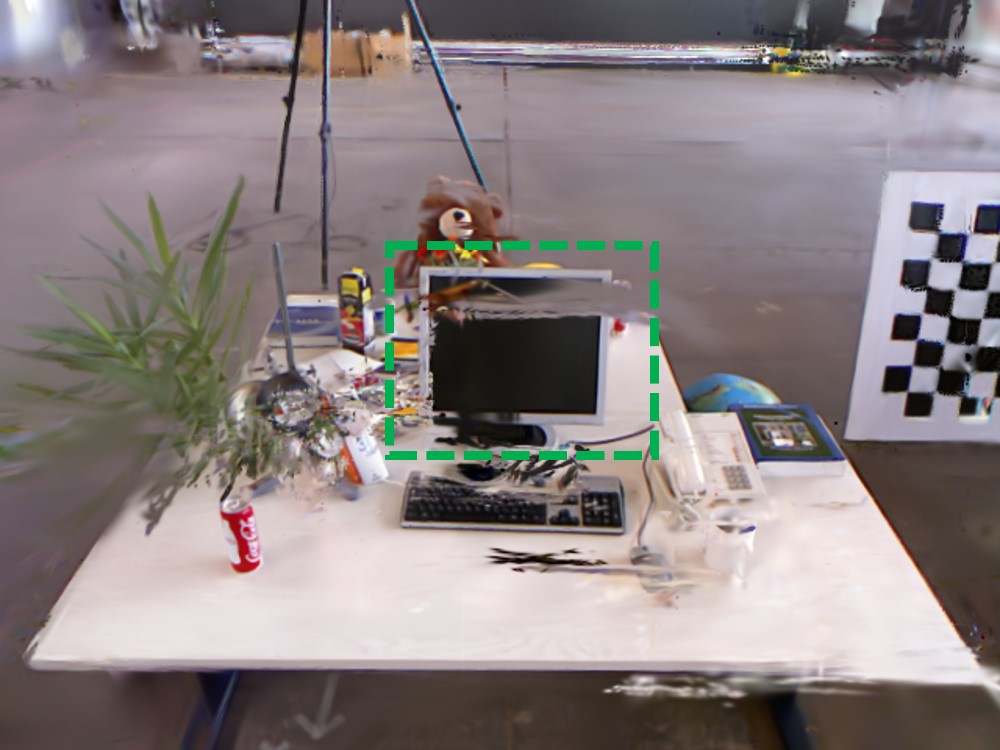}
        \\[0.1cm]
        \includegraphics[width=\linewidth]{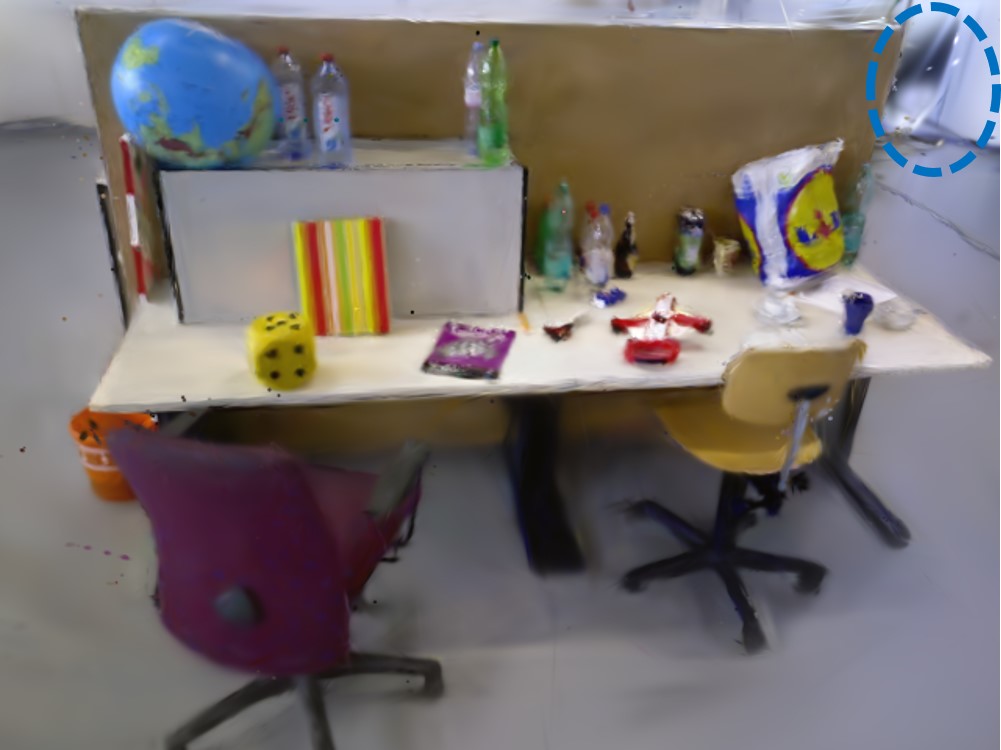}
        \\[-0.2cm]
        \end{minipage}
    }  
    \hspace{-0.4cm}
    \subfigure[\textbf{Ours}]{
        \begin{minipage}{\imw\linewidth}
        \includegraphics[width=\linewidth]{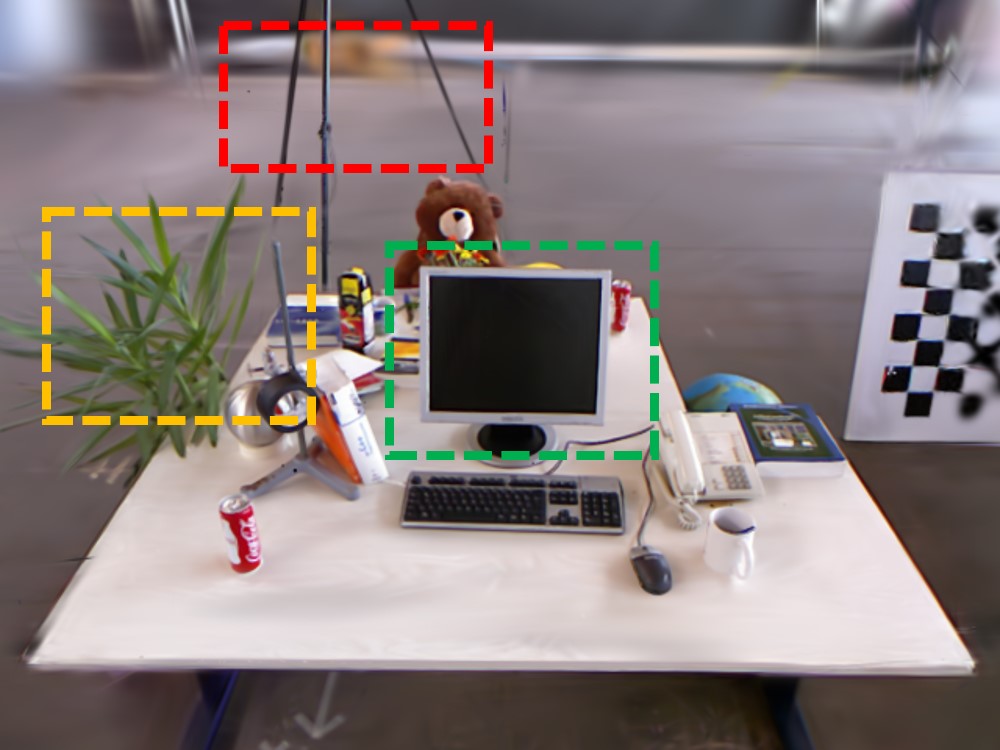}
        \\[0.1cm]
        \includegraphics[width=\linewidth]{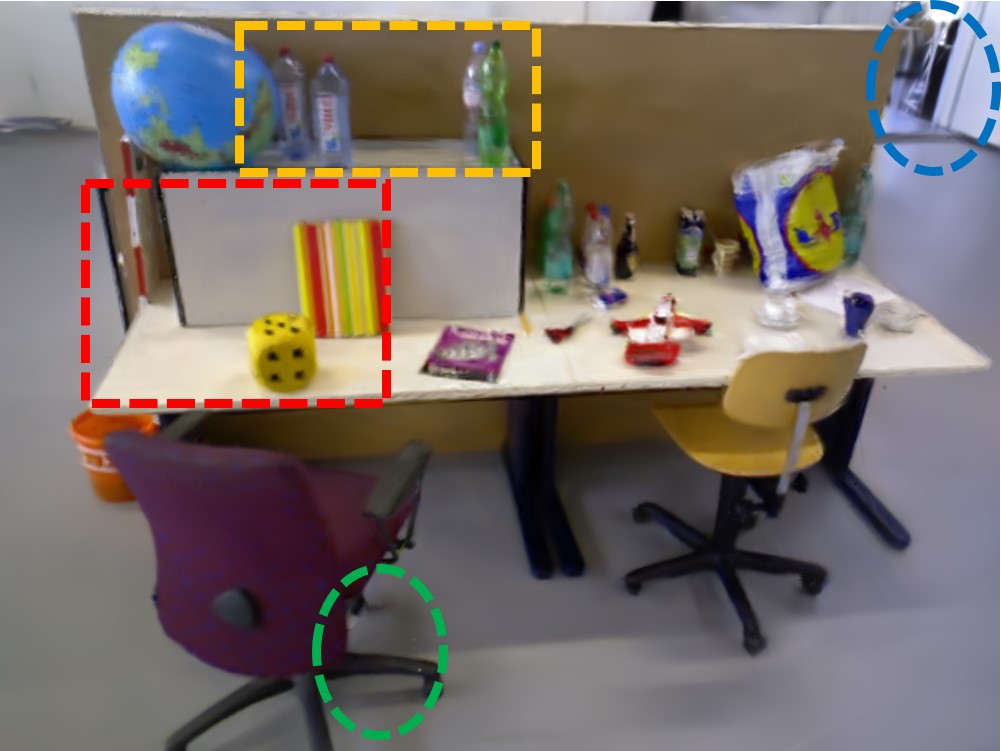}
        \\[-0.2cm]
        \end{minipage}
    }  
    \caption{Qualitative comparison of rendering performance on the TUM RGB-D.}
    \vspace{-1.2em}
    \label{fig:tum_rgbd}
\end{figure*}

\textbf{Novel View Rendering and Map Points Number.}
Although SplaTAM~\cite{Keetha_2024_CVPR} requires the longest optimization time, its rendering quality is not significantly better than that of other methods, except for the fr1-desk RGB-D sequence.
It is worth noting that we used the result from MonoGS~\cite{Matsuki_2024_CVPR} before performing color-refinement to ensure a fair comparison.
On the Replica dataset, GS-ICP SLAM~\cite{ha2024rgbdgsicpslam} often achieves better results because the dataset contains less noise and fewer outliers. However, on the TUM RGB-D dataset, its results are worse than ours, it can be seen from Fig.~\ref{fig:tum_rgbd} that the obvious differences.
Among them, we use different colors and shapes to highlight the differences in details between the compared baselines and our method. In these regions, our proposed method can achieve better and higher rendering results.
On the Replica monocular dataset (room2), the images rendered by our proposed method capture more fine details compared with MonoGS~\cite{Matsuki_2024_CVPR} and Photo-SLAM~\cite{hhuang2024photoslam}, as shown in Fig.~\ref{fig:replica_monocular}. We zoom in and out to highlight these differences more clearly.

Additionally, compared with Photo-SLAM~\cite{hhuang2024photoslam}, 
which relies on feature-point initialization, our method further improves the rendering quality (PSNR) by employing PG sampling to initialize 3D Gaussian primitives as we described in Sec.~\ref{subsec:Patch-Grid}. 
However, MonoGS~\cite{Matsuki_2024_CVPR} provides more accurate covisibility estimation, especially when dealing with occlusion, thereby improving the consistency of the reconstructed 3D Gaussian map on real-world datasets (fr2-xyz, fr3-office). Although it can generate the map with fewer points, as shown in Tab.~\ref{tab:tum_eval}, this also leads to a decrease in PSNR.

\textbf{Run-Time Analysis and Memory Usage.}
Among all the baseline methods, SplaTAM~\cite{Keetha_2024_CVPR} has the lowest frame rate (\textless 1 FPS), which depends on the number of iterations and the keyframe selection strategy. Even so, having more iterations in the optimization does not necessarily lead to better results. A detailed analysis of this aspect is provided in GS-ICP SLAM~\cite{ha2024rgbdgsicpslam}. 
Note that in~\cite{ha2024rgbdgsicpslam}, two versions of the tracking module are provided: one limited to 30 FPS and another without a speed limit. In our experiments, we observed that the the reconstruction quality of the latter is significantly lower than the former (\textasciitilde4dB or more). Without loss of generality, since both GS-ICP SLAM~\cite{ha2024rgbdgsicpslam} and Photo-SLAM~\cite{hhuang2024photoslam} can run at a speed of more than 30 FPS, we focus on comparing other more valuable and meaningful metrics.

Compared with these two methods~\cite{ha2024rgbdgsicpslam, hhuang2024photoslam}, our approach incorporates PG sampling points and merges 3D Gaussian primitives in voxels, as we analyzed in Sec.~\ref{subsec:voxel_based_merge}, the above main functions are all implemented on CUDA and are all in milliseconds. Therefore, our system can also reach more than 30 FPS and can still run in real-time and on edge devices.
As you can see in Tab.~\ref{tab:replica_eval_office}, ~\ref{tab:replica_eval_room} and Tab.~\ref{tab:tum_eval}, MonoGS~\cite{Matsuki_2024_CVPR} takes up the most GPU memory in most scenarios, since it shares the 3D Gaussian map of the back-end mapping process to the front-end tracking process via deep copy. Although it is the fast and usually taking 3 to 5 milliseconds, this directly leads to higher GPU memory usage compared to other methods.
It can be seen that on all datasets, our method consistently uses the least GPU memory, thanks to the voxel-space merging of 3D Gaussian primitives proposed in Sec.~\ref{subsec:voxel_based_merge}.

\subsection{{Real-World Experiments}}
\label{subsec:Real-World Experiments}
We deploy our system on a Jetson AGX Orin 64GB with an Intel RealSense D435i RGB-D camera. We have already obtained the intrinsic parameters of the camera. For keyframes detected by the tracking module of the front-end, we perform the PG-based initialization (Fig.~\ref{fig:initialization}), to achieve densification and model the entire scene (Fig.~\ref{fig:densification}). This step is used to initialize the 3D Gaussian primitives (Fig.~\ref{fig:3d_gaussian_mapping}) and improve the quality of the online reconstruction (Fig.~\ref{fig:online_reconstruction}).
\section{{Conclusion}}
\label{sec:conclusion}

We present a 3DGS-based real-time SLAM system in this work. 
In the system's front-end, we perform PG sampling on keyframes to better model the entire scene and improve image rendering quality. 
In the back-end, we merge geometrically similar 3D Gaussian primitives within the same voxel to reduce GPU memory usage without affecting runtime performance. 
Moreover, we conduct a real-world experiment on an edge device, demonstrating that our approach represents a step forward in real-time robotics applications.

\bibliographystyle{IEEEtran}
\bibliography{main}

\begin{thebibliography}{10}
\providecommand{\url}[1]{#1}
\csname url@rmstyle\endcsname
\providecommand{\newblock}{\relax}
\providecommand{\bibinfo}[2]{#2}
\providecommand\BIBentrySTDinterwordspacing{\spaceskip=0pt\relax}
\providecommand\BIBentryALTinterwordstretchfactor{4}
\providecommand\BIBentryALTinterwordspacing{\spaceskip=\fontdimen2\font plus
\BIBentryALTinterwordstretchfactor\fontdimen3\font minus \fontdimen4\font\relax}
\providecommand\BIBforeignlanguage[2]{{%
\expandafter\ifx\csname l@#1\endcsname\relax
\typeout{** WARNING: IEEEtran.bst: No hyphenation pattern has been}%
\typeout{** loaded for the language `#1'. Using the pattern for}%
\typeout{** the default language instead.}%
\else
\language=\csname l@#1\endcsname
\fi
#2}}

\bibitem{kinectfusion}
R.~A. Newcombe, S.~Izadi, O.~Hilliges, D.~Molyneaux, D.~Kim, A.~J. Davison, P.~Kohi, J.~Shotton, S.~Hodges, and A.~Fitzgibbon, ``Kinectfusion: Real-time dense surface mapping and tracking,'' in \emph{2011 10th IEEE international symposium on mixed and augmented reality}.\hskip 1em plus 0.5em minus 0.4em\relax Ieee, 2011, pp. 127--136.

\bibitem{dai2017bundlefusion}
A.~Dai, M.~Nie{\ss}ner, M.~Zollh{\"o}fer, S.~Izadi, and C.~Theobalt, ``Bundlefusion: Real-time globally consistent 3d reconstruction using on-the-fly surface reintegration,'' \emph{ACM Transactions on Graphics (ToG)}, vol.~36, no.~4, p.~1, 2017.

\bibitem{infinitam}
V.~A. Prisacariu, O.~Kahler, M.~M. Cheng, C.~Y. Ren, J.~Valentin, P.~H.~S. Torr, I.~D. Reid, and D.~W. Murray, ``{A Framework for the Volumetric Integration of Depth Images},'' \emph{ArXiv e-prints}, 2014.

\bibitem{6599048}
M.~Keller, D.~Lefloch, M.~Lambers, S.~Izadi, T.~Weyrich, and A.~Kolb, ``Real-time 3d reconstruction in dynamic scenes using point-based fusion,'' in \emph{2013 International Conference on 3D Vision - 3DV 2013}, 2013, pp. 1--8.

\bibitem{8794101}
K.~Wang, F.~Gao, and S.~Shen, ``Real-time scalable dense surfel mapping,'' in \emph{2019 International Conference on Robotics and Automation (ICRA)}, 2019, pp. 6919--6925.

\bibitem{nerf}
B.~Mildenhall, P.~P. Srinivasan, M.~Tancik, J.~T. Barron, R.~Ramamoorthi, and R.~Ng, ``Nerf: Representing scenes as neural radiance fields for view synthesis,'' in \emph{European conference on computer vision}.\hskip 1em plus 0.5em minus 0.4em\relax Springer, 2020, pp. 405--421.

\bibitem{kerbl3Dgaussians}
B.~Kerbl, G.~Kopanas, T.~Leimk{\"u}hler, and G.~Drettakis, ``3d gaussian splatting for real-time radiance field rendering,'' \emph{ACM Transactions on Graphics}, vol.~42, no.~4, July 2023.

\bibitem{pointslam}
E.~Sandstr{\"o}m, Y.~Li, L.~Van~Gool, and M.~R. Oswald, ``Point-slam: Dense neural point cloud-based slam,'' in \emph{Proceedings of the IEEE/CVF International Conference on Computer Vision}, 2023, pp. 18\,433--18\,444.

\bibitem{sucar2021imap}
E.~Sucar, S.~Liu, J.~Ortiz, and A.~J. Davison, ``imap: Implicit mapping and positioning in real-time,'' in \emph{Proceedings of the IEEE/CVF International Conference on Computer Vision}, 2021, pp. 6229--6238.

\bibitem{zhu2022nice}
Z.~Zhu, S.~Peng, V.~Larsson, W.~Xu, H.~Bao, Z.~Cui, M.~R. Oswald, and M.~Pollefeys, ``Nice-slam: Neural implicit scalable encoding for slam,'' in \emph{Proceedings of the IEEE/CVF Conference on Computer Vision and Pattern Recognition}, 2022, pp. 12\,786--12\,796.

\bibitem{rosinol2022nerf}
A.~Rosinol, J.~J. Leonard, and L.~Carlone, ``Nerf-slam: Real-time dense monocular slam with neural radiance fields,'' \emph{arXiv preprint arXiv:2210.13641}, 2022.

\bibitem{maggio2022loc}
D.~Maggio, M.~Abate, J.~Shi, C.~Mario, and L.~Carlone, ``Loc-nerf: Monte carlo localization using neural radiance fields,'' \emph{arXiv preprint arXiv:2209.09050}, 2022.

\bibitem{Keetha_2024_CVPR}
N.~Keetha, J.~Karhade, K.~M. Jatavallabhula, G.~Yang, S.~Scherer, D.~Ramanan, and J.~Luiten, ``Splatam: Splat track \& map 3d gaussians for dense rgb-d slam,'' in \emph{Proceedings of the IEEE/CVF Conference on Computer Vision and Pattern Recognition (CVPR)}, June 2024, pp. 21\,357--21\,366.

\bibitem{Matsuki_2024_CVPR}
H.~Matsuki, R.~Murai, P.~H. Kelly, and A.~J. Davison, ``Gaussian splatting slam,'' in \emph{Proceedings of the IEEE/CVF Conference on Computer Vision and Pattern Recognition (CVPR)}, June 2024, pp. 18\,039--18\,048.

\bibitem{ha2024rgbdgsicpslam}
S.~Ha, J.~Yeon, and H.~Yu, ``Rgbd gs-icp slam,'' in \emph{European Conference on Computer Vision (ECCV)}, 2024.

\bibitem{hhuang2024photoslam}
H.~Huang, L.~Li, C.~Hui, and S.-K. Yeung, ``Photo-slam: Real-time simultaneous localization and photorealistic mapping for monocular, stereo, and rgb-d cameras,'' in \emph{Proceedings of the IEEE/CVF Conference on Computer Vision and Pattern Recognition}, 2024.

\bibitem{peng2024rtgslam}
Z.~Peng, T.~Shao, L.~Yong, J.~Zhou, Y.~Yang, J.~Wang, and K.~Zhou, ``Rtg-slam: Real-time 3d reconstruction at scale using gaussian splatting,'' in \emph{ACM SIGGRAPH Conference Proceedings, Denver, CO, United States, July 28 - August 1, 2024}, 2024.

\bibitem{eslam}
M.~M. Johari, C.~Carta, and F.~Fleuret, ``Eslam: Efficient dense slam system based on hybrid representation of signed distance fields,'' in \emph{Proceedings of the IEEE/CVF Conference on Computer Vision and Pattern Recognition}, 2023, pp. 17\,408--17\,419.

\bibitem{coslam}
H.~Wang, J.~Wang, and L.~Agapito, ``Co-slam: Joint coordinate and sparse parametric encodings for neural real-time slam,'' in \emph{Proceedings of the IEEE/CVF Conference on Computer Vision and Pattern Recognition}, 2023, pp. 13\,293--13\,302.

\bibitem{muller2022instant}
T.~M{\"u}ller, A.~Evans, C.~Schied, and A.~Keller, ``Instant neural graphics primitives with a multiresolution hash encoding,'' \emph{ACM Transactions on Graphics (ToG)}, vol.~41, no.~4, pp. 1--15, 2022.

\bibitem{TensoRF}
A.~Chen, Z.~Xu, A.~Geiger, J.~Yu, and H.~Su, ``Tensorf: Tensorial radiance fields,'' in \emph{European Conference on Computer Vision (ECCV)}, 2022.

\bibitem{MurArtal15tro}
R.~Mur-Artal, J.~M.~M. Montiel, and J.~D. Tard{\'o}s, ``{ORB-SLAM}: a versatile and accurate monocular {SLAM} system,'' \emph{{IEEE} Trans. Robot.}, vol.~31, no.~5, pp. 1147--1163, 2015.

\bibitem{campos2021orb3}
C.~Campos, R.~Elvira, J.~J.~G. Rodr{\'\i}guez, J.~M. Montiel, and J.~D. Tard{\'o}s, ``Orb-slam3: An accurate open-source library for visual, visual--inertial, and multimap slam,'' \emph{IEEE Transactions on Robotics}, vol.~37, no.~6, pp. 1874--1890, 2021.

\bibitem{7898369}
J.~Engel, V.~Koltun, and D.~Cremers, ``Direct sparse odometry,'' \emph{IEEE Transactions on Pattern Analysis and Machine Intelligence}, vol.~40, no.~3, pp. 611--625, 2018.

\bibitem{8593376}
X.~Gao, R.~Wang, N.~Demmel, and D.~Cremers, ``Ldso: Direct sparse odometry with loop closure,'' in \emph{2018 IEEE/RSJ International Conference on Intelligent Robots and Systems (IROS)}, 2018, pp. 2198--2204.

\bibitem{9546534}
Z.~Yuan, K.~Cheng, J.~Tang, and X.~Yang, ``Rgb-d dso: Direct sparse odometry with rgb-d cameras for indoor scenes,'' \emph{IEEE Transactions on Multimedia}, vol.~24, pp. 4092--4101, 2022.

\bibitem{koide2021voxelized}
K.~Koide, M.~Yokozuka, S.~Oishi, and A.~Banno, ``Voxelized gicp for fast and accurate 3d point cloud registration,'' in \emph{2021 IEEE International Conference on Robotics and Automation (ICRA)}.\hskip 1em plus 0.5em minus 0.4em\relax IEEE, 2021, pp. 11\,054--11\,059.

\bibitem{icp}
P.~J. Besl and N.~D. McKay, ``Method for registration of 3-d shapes,'' in \emph{Sensor fusion IV: control paradigms and data structures}, vol. 1611.\hskip 1em plus 0.5em minus 0.4em\relax Spie, 1992, pp. 586--606.

\bibitem{Lee_2024_CVPR}
J.~C. Lee, D.~Rho, X.~Sun, J.~H. Ko, and E.~Park, ``Compact 3d gaussian representation for radiance field,'' in \emph{Proceedings of the IEEE/CVF Conference on Computer Vision and Pattern Recognition (CVPR)}, 2024, pp. 21\,719--21\,728.

\bibitem{Kuemmerle11icra}
R.~K{\"u}mmerle, G.~Grisetti, H.~Strasdat, K.~Konolige, and W.~Burgard, ``g2o: A general framework for graph optimization,'' in \emph{{IEEE} Int. Conf. Robot. Autom. (ICRA)}, 2011.

\bibitem{Dellaert12tr}
F.~Dellaert, ``Factor graphs and {GTSAM}: A hands-on introduction,'' Georgia Institute of Technology, Tech. Rep. GT-RIM-CP\&R-2012-002, Sept. 2012.

\bibitem{givens1984class}
C.~R. Givens and R.~M. Shortt, ``A class of wasserstein metrics for probability distributions.'' \emph{Michigan Mathematical Journal}, vol.~31, no.~2, pp. 231--240, 1984.

\bibitem{dataset2019replica}
J.~Straub, T.~Whelan, L.~Ma, Y.~Chen, E.~Wijmans, S.~Green, J.~J. Engel, R.~Mur-Artal, C.~Ren, S.~Verma, A.~Clarkson, M.~Yan, B.~Budge, Y.~Yan, X.~Pan, J.~Yon, Y.~Zou, K.~Leon, N.~Carter, J.~Briales, T.~Gillingham, E.~Mueggler, L.~Pesqueira, M.~Savva, D.~Batra, H.~M. Strasdat, R.~D. Nardi, M.~Goesele, S.~Lovegrove, and R.~Newcombe, ``The {R}eplica dataset: A digital replica of indoor spaces,'' \emph{arXiv preprint arXiv:1906.05797}, 2019.

\bibitem{dataset2012tum}
J.~Sturm, N.~Engelhard, F.~Endres, W.~Burgard, and D.~Cremers, ``A benchmark for the evaluation of rgb-d slam systems,'' in \emph{Proc. of the International Conference on Intelligent Robot Systems (IROS)}, Oct. 2012.

\end{thebibliography}
\end{document}